\documentclass[12pt]{article}

\usepackage{scicite}
\usepackage[]{graphicx}
\usepackage{times}
\usepackage{xcolor}
\usepackage{amsmath}
\usepackage{siunitx}
\usepackage{algorithm}
\usepackage{algpseudocode}
\usepackage[labelfont=bf]{caption}

\newcommand\rev[1]{\textcolor{black}{#1}}  
\usepackage{titlesec}
\usepackage{titling}
\titleformat{\subparagraph}
  {\normalfont\rmfamily\itshape\bfseries}{\thesubparagraph}{}{}

\newenvironment{sciabstract}{%
\begin{quote} \bf}
{\end{quote}}

\title{\rev{Locomotion as Manipulation with ReachBot  }}

\author
{Tony G. Chen$^{1\ast}$, Stephanie Newdick$^{2}$, Julia Di$^{1}$, Carlo Bosio$^{1}$, Nitin Ongole$^{2}$,\\ Mathieu Lap\^{o}tre$^{3}$, Marco Pavone$^{2}$, Mark R. Cutkosky$^{1}$\\
\\
\normalsize{$^{1}$Dept. of Mechanical Engineering, $^{2}$Dept. of Aeronautics and Astronautics,}\\
\normalsize{$^{3}$Dept. of Earth and Planetary Sciences, Stanford University,}\\
\normalsize{424 Panama Mall, Stanford 94305, USA}\\
\normalsize{$^\ast$Corresponding author:  agchen@stanford.edu.}
}

\date{}

\begin{document} 




\maketitle


\begin{sciabstract}

Caves and lava tubes on the Moon and Mars are sites of geological and astrobiological interest but consist of terrain that is inaccessible with traditional robot locomotion.
To support the exploration of these sites, we present ReachBot, a robot that uses extendable booms as appendages to manipulate itself with respect to irregular rock surfaces. The booms terminate in grippers equipped with microspines and provide ReachBot with a large workspace, allowing it to achieve force closure in enclosed spaces such as the walls of a lava tube.  To propel ReachBot, we present a contact-before-motion planner for non-gaited legged locomotion that utilizes internal force control, similar to a multi-fingered hand, to keep its long, slender booms in tension. Motion planning also depends on finding and executing secure grips on rock features. We use a Monte Carlo simulation to inform gripper design and predict grasp strength and variability. Additionally, we use a two-step perception system to identify possible grasp locations. To validate our approach and mechanisms under realistic conditions, we deployed a single ReachBot arm and gripper in a lava tube in the Mojave Desert. The field test confirmed that ReachBot will find many targets for secure grasps with the proposed kinematic design. 
\\
\\
We present grasp and motion plans and field test results for a robot that uses extending booms to navigate difficult terrain. 
\end{sciabstract}


\section*{Introduction} \label{sec:Introduction}

\begin{figure}[h]
    \centering
    \includegraphics[width=1\textwidth]{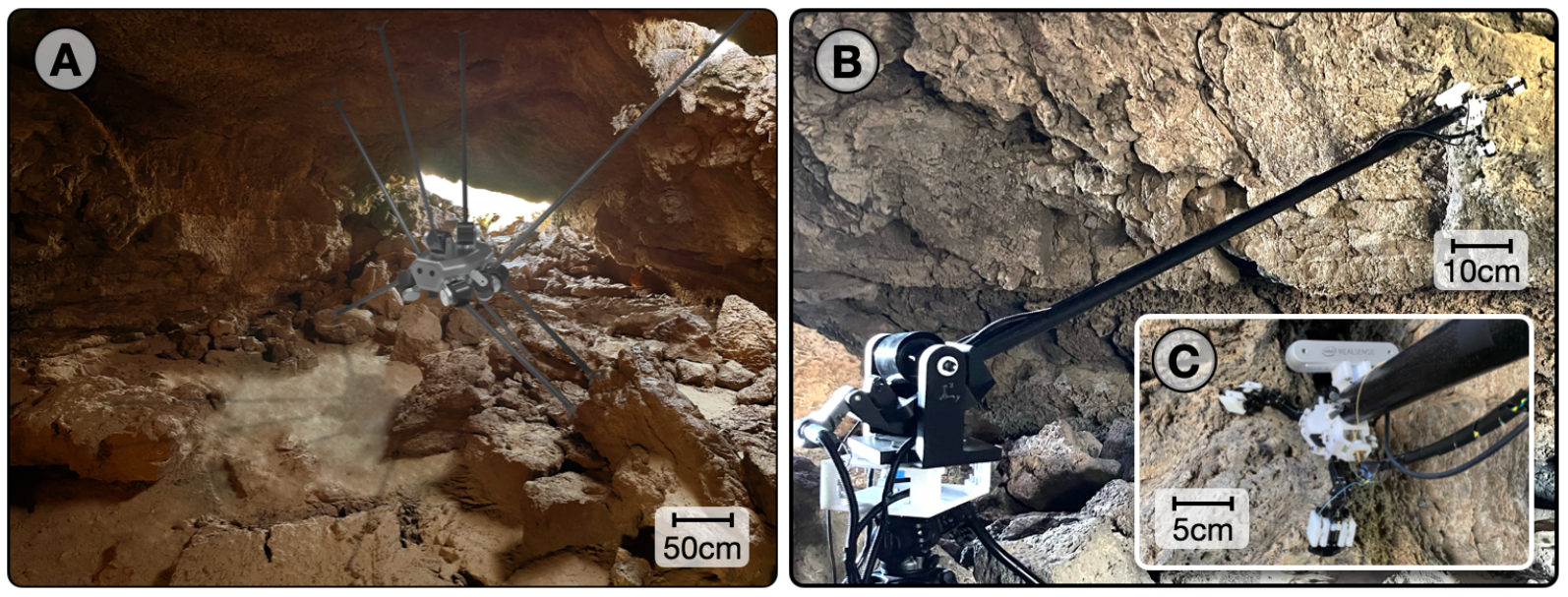}
    \caption{\textbf{ReachBot exploring an analog martian lava tube.} (A) shows a rendering of a full ReachBot configuration overlaid on our field test site in the Lavic Lake volcanic field in the Mojave Desert. (B) shows the single-boom prototype in the same lava tube to demonstrate deployer, perception system, and microspine gripper, which can be seen grasping from another perspective in (C).
    }
    \label{fig:glamour}
\end{figure}

Cliffs, caves, and lava tubes on the moon and Mars have been identified as locations of intense geological and astrobiological interest. Due to their insulating and shielding properties, these caverns provide relatively stable conditions which may promote mineral precipitation and microbial growth~\cite{LeveilleDatta2010}. On Mars, these ancient subsurface environments are nearly unchanged from when the martian surface was potentially habitable while life on Earth was already evolving ~\cite{OnstottEhlmannEtAl2019, GrotzingerSumnerEtAl2013, OhtomoKakegawaEtAl2014}. 
Additionally, sheltered cavern environments could provide sites for future human habitation~\cite{Boston2010}.
There is therefore a growing interest in exploring cave-like features that have so far only been examined via orbiter-based remote sensing. Such planetary operations require robots capable of mobility and manipulation in a variety of terrains, particularly rocky terrain and vertical or overhanging surfaces where anchor points may be sparse. NASA's interest in these sites motivates robots that combine sparse-anchored mobility with high-wrench (force and moment) manipulation. However, there is a key technology gap in existing solutions: small robots typically have a small reach and limited wrench capability. Conversely, large robots, particularly those that employ articulated serial links, are hampered by high mass and complexity that scale poorly with increased reach.

\rev{ReachBot is a robot concept} that locomotes by grasping rock features with multiple appendages, traversing environments that require climbing even when anchor points are sparse (Fig.~\ref{fig:glamour}). 
Using lightweight, extendable booms as appendages, ReachBot achieves a form factor that resembles a
harvestman arachnid (\emph{Opiliones}) \cite{spagna2012terrestrial}: a small body with very long limbs.
Extendable booms have been developed for space applications such as antenna structures \cite{Fernandez2017, FootdaleMurphey2014, SpenceWhiteEtAl2018} because they are light and compact when rolled up, but are strong---especially in tension---when deployed and capable of extending many times the span of the robot body. 
For ReachBot, the tips of the booms are equipped with pivoting wrists and grippers that use arrays of microspines to grasp rocky surfaces. As we discuss in the following sections, this arrangement allows ReachBot to move by manipulating its body with respect to the terrain. 

Locomotion as manipulation provides access to the floor, walls, and ceilings of caves or lava tubes. \rev{To increase scientific value, robotic investigation of these systems should include subsurface stratigraphy of the walls, which may reveal a great deal about the geologic history of the planetary body~\cite{BaioniSgavetti2013}. }
More broadly, 
vertical outcrops (whether in caves or along cliffs) offer a unique opportunity to read the geologic record in a spatially continuous fashion, but can only be accessed by climbing.
These interesting geological features can also contain large crevasses and loose rubble piles that make them difficult for traditional robots to navigate.
Existing legged and climbing robots could access some sites of interest, but the lengths of their limbs restrict them to movement along one surface, neglecting the advantage offered by terrain features on the surrounding walls of a cave or lava tube. Conversely, a robot that
manipulates itself with extendable booms can leverage anchor points in all directions to bypass obstacles and access vertical outcrops, as well as assume a wide variety of possible configurations, bracing stances, and force application options.

In previous publications, we introduced the ReachBot concept and discussed planar motion planning and control \cite{SchneiderBylardEtAl2022, ChenMillerEtAl2022, NewdickChenEtAl2023, NewdickOngoleEtAl2023}.
In this paper, we extend the analysis and motion planning to fully three-dimensional scenarios and report on field tests of ReachBot technology to pave the way for full-scale ReachBot deployment.

\rev{Although locomotion via extending booms is particular to ReachBot, the kinematics and dynamics share similarities to some other established areas of work.}
The model used for motion planning and control adapts concepts from legged locomotion and from manipulation with multifingered hands, including the formation of a grasp matrix and the consideration of internal and external grasp forces and force closure ~\cite{BicchiKumar2000,HanTrinkleEtAl2000,LiHsuEtAl1989,NagaiYoshikawa1993}. We use a decoupled motion planning strategy that combines discrete footstep planning with continuous trajectory optimization. Similar strategies have been successful for legged, climbing\rev{, and brachiating} robots~\cite{KuindersmaDeitsEtAl2015,HauserBretlEtAl2005,bretl2006motion,parness2017lemur, nakanishi2000brachiating}, but our combination of a large workspace and potentially sparse anchor points requires a more versatile motion plan that does not rely on a predefined gait. \rev{In addition, we assume that ReachBot moves slowly enough that it does not have inertial swinging dynamics.}
In particular, we look to contact-before-motion planners, which offer solutions for robots in terrain that require careful footstep planning~\cite{Parness2017a,ZhangLatomb2013}.
ReachBot requires a relatively high effort for each footstep: it must target a site, extend a boom, grasp, and tension the boom. Grasp failures can also degrade the gripper spines and send large transient forces into the robot structure. Accordingly, we seek to minimize grasp failures via risk-aware motion planning, requiring a model of grasp strength. 
\rev{Finally, ReachBot also shares similarities with cable-based robots \cite{lamaury2012design,capua2014spiderbot,qian2018review}, including some cable-based designs proposed for planetary exploration or construction activities \cite{seriani2016modular, mcgarey2018developing, nesnas2012axel}. Like cables, the booms exert forces primarily along their long axes and similar issues arise with respect to computing the stability of a stance. Unlike cables, however, booms can be steered toward targets.}

Grasping or climbing rocky surfaces with spines has been addressed in various publications, including \rev{some that address space applications} \cite{asbeck2006scaling,parness2017lemur,wang2019spinyhand,zi2023mechanical,ParnessFrostEtAl2013,backus2020design,ParnessWilligEtAl2017}. The analysis of collections of spines has also been addressed in detail \cite{asbeck2012designing,wang2019spinyhand,jiang2018stochastic,iacoponi2020simulation}.
We draw upon this work for the design and analysis of ReachBot's grippers. \rev{ReachBot targets convex features with a radius of curvature approximately on the scale of its grippers. This choice reduces the density of targets but results in grasps with a high ratio of pull-off to grasp force. It also requires ReachBot to identify such features in the surrounding terrain.}

Our approach to grasp site identification builds upon extensive work on perceiving outdoor terrain for legged and climbing robots. This includes offline methods of pre-scanning a terrain map \cite{zucker2010optimization, neuhaus2011comprehensive}, and online methods that continuously construct and re-plan based on onboard sensors \cite{kolter2009stereo, tranzatto2022cerberus}. Because ReachBot must be able to navigate previously unknown caves, onboard sensing is crucial for identifying grasp sites for the planner. Existing microspine-based climbing robots have a comparatively small workspace, so their perception systems can construct detailed surface scans to enable grasp site prediction \cite{parness2017lemur}. Because ReachBot has a large workspace, we present a two-stage (far and near) \rev{vision-based} perception strategy to identify rock features and estimate their ability to provide a strong grasp.

\rev{We present field tests with a robot that locomotes by manipulating itself in caves and similar terrain.} The technical contributions include: a three-dimensional simulation and motion planner for ReachBot, identifying gaps in existing work on locomotion in cave-like environments; a kinematic model of a general three-finger gripper equipped with spines on a convex rock surface, used to compute probabilistic force limit surfaces for grasp planning; a lightweight underactuated gripper specialized for secure grasps on irregular convex surfaces; a perception pipeline that identifies graspable sites from a distance; empirical field data to support the grasp modeling, grasp site identification, and gripper design. The tests were conducted in a lava tube in the Mojave Desert as a proxy for lava tubes on Mars.

\section*{Results} \label{sec:Results}

\subsection*{ReachBot model, control, and motion planning}

\begin{figure}[h!]
    \centering
    \includegraphics[width=0.9\textwidth]{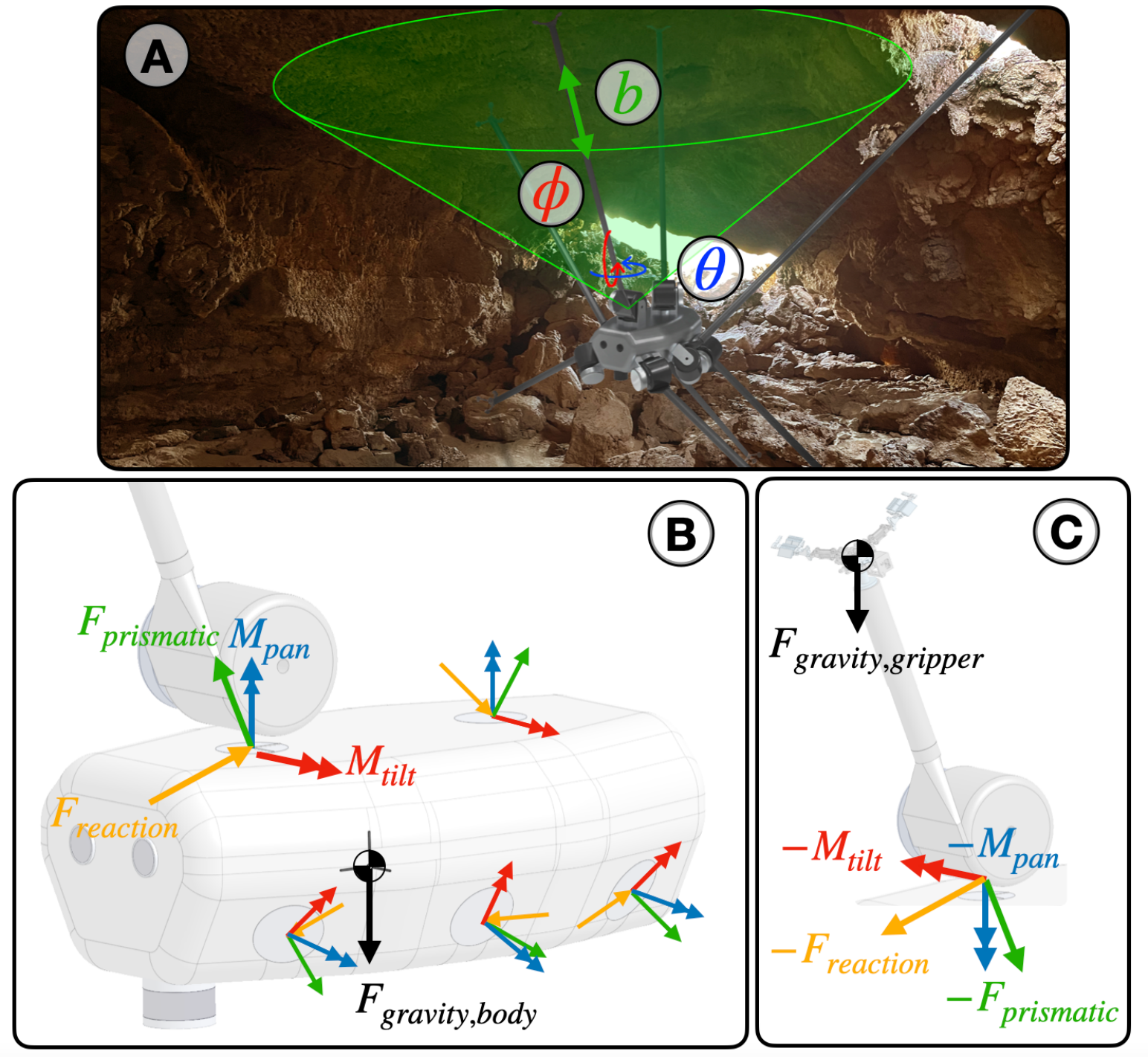}
    \caption{\textbf{Visualization of ReachBot's kinematic workspace.} (A) Each shoulder joint has a pan, $\theta$, and tilt, $\phi$, range of motion in addition to a prismatic boom with extension $b$.
    The model treats boom deployers as points, with their mass incorporated into the mass of the robot body. A free-body diagram for ReachBot's body (B) includes the actuated pan moment $M_{\text{pan}}$, tilt moment $M_{\text{tilt}}$, and prismatic force $F_{\text{prismatic}}$ \rev{along the boom}, as well as a reaction force $F_{\text{reaction}}$ \rev{in the plane perpendicular to $F_{\text{prismatic}}$}, at each shoulder joint. \rev{There is no moment about the boom due to the free wrist.} Gravity ($F_{\text{gravity,body}} = m_\text{body}G$) acts at the center of mass. When a gripper is attached, it supports its own weight, but when detached, the shoulder joint must support it. The free-body diagram for an unattached gripper (C) includes interaction forces at the shoulder joint and gravity ($F_{\text{gravity,gripper}} = m_\text{gripper}G$). } 
    \label{fig:kinematics}
\end{figure}

The architecture for ReachBot used throughout this paper was developed as part of a system design trade study for a mission to a martian lava tube~\cite{NewdickChenEtAl2023}. This configuration of ReachBot has eight extendable booms --- three on either side and two overhead --- each with separate pan, tilt, and prismatic actuated joints. The range of motion of an example boom is shown in Fig.~\ref{fig:kinematics}(A). The gripper is mounted on a ball joint wrist with a single degree of actuation for grasping and releasing at the end of each boom.

ReachBot's long, slender booms are susceptible to buckling from compression or bending loads. We can respect their limitations by employing concepts used in multi-fingered dexterous manipulation, in particular force closure and internal force control. In dexterous manipulation, fingers apply unisense contact forces to manipulate an object~\cite{MurrayLiEtAl1994,BicchiKumar2000,HanTrinkleEtAl2000}. ReachBot moves analogously to a manipulated object. In place of fingers that push, ReachBot has booms that pull. By ensuring that the booms are always in tension, we leverage the tensile strength of the booms and maintain the stiffness of the entire structure, analogous to a bicycle wheel with tensioned spokes.

ReachBot executes two types of continuous movement that mirror manipulation with a multi-fingered hand. In body movement (``in-grasp manipulation"), all grippers are attached to the environment, and in end-effector movement (``finger gaiting" or ``re-grasp manipulation"),  a gripper detaches and moves to a new grasp point.
The free-body diagrams in Fig.~\ref{fig:kinematics}(B,C) represent the dynamics of ReachBot's body and a detached gripper under the assumption that the distributed mass of each boom is small enough that we can approximate the inertial dynamics using a lumped parameter model consisting of the masses of the body and the grippers. 
We assume that the attached grippers do not move from their anchor points so the control inputs shown in Fig.~\ref{fig:kinematics}(B) apply a wrench directly to the surroundings. Although the attached grippers support themselves, ReachBot must support any detached grippers. In general, we decouple body movement and end-effector movement as two modes of locomotion with distinct governing dynamics.

During both types of movement, we define a Jacobian, $\mathbf{J}$, that relates joint velocities to body linear and angular velocities.  
We can apply a wrench on the robot body using the relationship
\begin{equation}
    W = -\mathbf{J}^T \tau,
    \label{eq:w=Jtau}
\end{equation}
where $W$ is the applied wrench on the robot body and $\tau$ is a vector of control inputs (one prismatic force and two rotational torques for each shoulder joint). We can solve this equation for $\tau$, but we must apply an additional constraint to maintain tension in the booms. Because there is redundant actuation in the system, $\mathbf{J}$ is not square; therefore, we define
\begin{equation}
    \hat{\tau} = \tau + C\mathcal{N}(\mathbf{J}^T)
    \label{eq:tauhat}
\end{equation}
where $\mathcal{N}(\mathbf{J}^T)$ is the null space of $\mathbf{J}^T$ and $C$ is a vector of dimension$(\mathcal{N}(\mathbf{J}^T))$ scalars. 
We use linear optimization to find a solution $C$ to constrain all prismatic forces to be in tension and minimize control effort. The resulting torque, $\hat{\tau}$, is applied to stabilization, control, or trajectory tracking.

\begin{figure}[htp]
    \centering \includegraphics[width=0.9\textwidth]{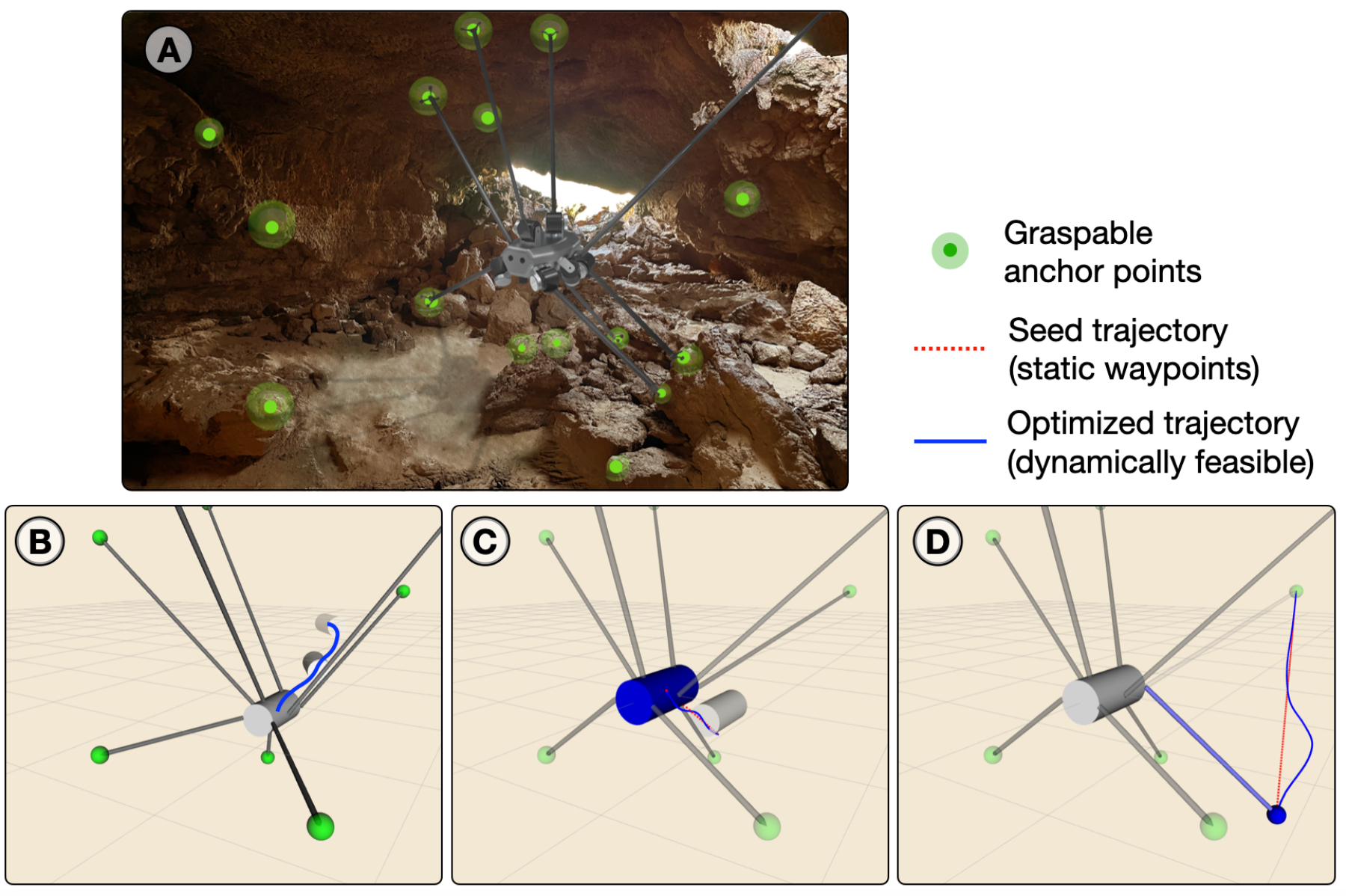}
    \caption{\textbf{Motion planning simulation based on a lava tube.} (A) shows a rendering of ReachBot in a lava tube near Mojave, CA with graspable anchor points identified with green highlighting. (B) shows a sequence of body positions corresponding to vertices in the footstep plan graph. (C) shows a body motion during which all eight grippers remain attached. (D) shows an end effector motion in which a gripper is detached and reattached at a new location. In each continuous phase, the seed trajectory (sequence of static waypoints) is shown as a dotted red line. The dynamically feasible optimized trajectory is shown as a solid blue curve.}
    \label{fig:motion_planning}
\end{figure}

Motion planning for ReachBot requires a combination of discrete footstep planning and continuous trajectory optimization. To take advantage of ReachBot's large workspace and accommodate the highly irregular terrain, we introduce a decoupled contact-before-motion plan wherein ReachBot uses discrete optimization to generate a footstep plan, then uses waypoint planning jointly with Sequential Convex Programming (SCP) to plan a sequence of continuous body and end-effector trajectories that carry out the footstep plan. The details of the discrete optimization are as per \cite{NewdickOngoleEtAl2023} but extended to 3D. 
\rev{We construct the simulated environment shown in Fig.~\ref{fig:motion_planning} based on anchor point data obtained in the Mojave Desert lava tube.}

The result of footstep planning is a sequence of stances (assignments that pair $n$ end-effectors to anchor points) linked by feasible transition poses with an overall goal of moving the body from a starting position to a goal. A feasible pose is one where the robot's configuration is not in collision with itself or with the environment and can hold itself statically within motor torque limits. We assume a transition between 8-stances is feasible if there exists a pose that is feasible: in the starting 8-stance; in the common 7-stance while holding the free boom at its starting anchor point; in the common 7-stance while holding the free boom at its ending anchor point; in the goal 8-stance. \rev{This assumption will be true except in pathological cases (for example abrupt, large changes in tunnel diameter) which are uncharacteristic of lava tubes.} To generate the plan, we construct a graph wherein each 8-stance is a vertex and each edge represents a feasible transition between stances. Fig.~\ref{fig:motion_planning}(B) shows a sequence of vertices in the footstep plan, with Fig.~\ref{fig:motion_planning}(C,D) showing examples of a body movement and end-effector movement phase, respectively, that ReachBot takes during its path from start to finish.

During body movement, ReachBot moves within an 8-stance to prepare for the next stance transition. The footstep path guarantees that each pose at the beginning and end of a required body movement is feasible, but we further require that every intermediate pose be feasible. To this end, we generate a discretized trajectory of
feasible poses by sampling locally from a straight-line trajectory, verifying at each point that a combination of forces exists to maintain that pose. We generate control inputs for each waypoint in the trajectory using Eq.~\eqref{eq:tauhat}. 
We then use this trajectory to warm-start the SCP solver, which minimizes control effort and generates a dynamically feasible trajectory. SCP sequentially approximates the non-convex body movement
problem with the following convex optimization problem
\begin{alignat}{3}
    \label{eq:opt_control}
& \min_{\substack{S_k,\ \tau_k \\ \forall k \in \{1 .. n_\text{timesteps}\}}} \quad \sum_{k=1}^{n_\text{timesteps}} \bigg( \quad && c(\tau_k) +
 \lambda_s ||S_k^{(w)} - S_k^{(w-1)}|| +
 \lambda_\tau ||\tau_k^{(w)} - \tau_k^{(w-1)}|| \bigg)
\end{alignat}
\vspace{5pt}
\begin{alignat*}{3}
&\quad \textrm{s.t.} \quad  S_{k+1} = A(S_k, \tau_k) & \quad \forall k \in \{1..n_\text{timesteps}-1\} \nonumber \\
&\quad \quad  S_1 = S_\text{start}  \nonumber \\
&\quad \quad  S_n = S_\text{goal} \nonumber \\
&\quad \quad \Theta_\text{min} \le \Theta_{k} \le \Theta_\text{max} & \quad \forall k \in \{1..n_\text{timesteps}\} \nonumber \\
&\quad \quad \tau_\text{min} \le \tau_{k} \le \tau_\text{max} &\quad \forall k \in \{1..n_\text{timesteps}\} \nonumber \\
\end{alignat*}
where the optimization variables are the state vector
\begin{equation}
    \label{eq:state_vector}
    S_k = [X_k, Q_k, P_k, L_k],
\end{equation}
which defines the body position $X = [x, y, z]$, orientation in quaternions $Q = [q_x, q_y, q_z, q_w]$, linear momentum $P = m_\text{body}[\dot{x}, \dot{y}, \dot{z}]$ and angular momentum $L = \mathbf{I_\text{body}}[\omega_x, \omega_y, \omega_z]$ at timestep $k$, and the control vector
\begin{equation}
    \label{eq:ctrl_vector}
    \tau_k = [F_{\text{prismatic},k\{i\}}, M_{\text{pan},{k\{i\}}},  M_{\text{tilt},{k\{i\}}} \quad \forall i \in \{1..n_\text{booms}\}],
\end{equation}
which defines the prismatic force, pan moment, and tilt moment applied to boom $i$ at timestep $k$. The control inputs $\tau_k$ are bounded by motor constraints based on hardware limitations. Similarly, joint positions
\begin{equation}
    \label{eq:joint_vector}
    \Theta_k = [b_{\text{prismatic},k\{i\}}, \theta_{\text{pan},{k\{i\}}},  \phi_{\text{tilt},{k\{i\}}} \quad \forall i \in \{1..n_\text{booms}\}]
\end{equation}
are bounded by kinematic constraints that limit the length of the boom, pan angle, and tilt angle. We use $(S_k)^{(w)}$ and $(\tau_{k})^{(w)}$ to denote the value of $(S_k)$ and $(\tau_{k})$ at SCP iteration $w$; where not explicitly noted all variables and derived quantities correspond to SCP iteration $w$.

\rev{To minimize control effort, we minimize $c(\tau_k)$, a weighted metric of forces and moments, at all timesteps. In particular, the weight of moments in this objective function is scaled with ReachBot's characteristic length, in this case the length of the body. For simplicity, we let $c(\tau_k) = ||\tau_k||$ in this work.}
In each iteration of the SCP, we penalize divergence from the local convexification using trust parameters for the state $\lambda_s$ and control $\lambda_\tau$.

The function $A$ represents ReachBot's dynamics linearized around the previous SCP iteration. Because the seed trajectory $(S^{(0)}, \tau^{(0)})$ is constructed from a string of static intermediate poses, this constraint is necessary to ensure the dynamic feasibility of the final trajectory. Finally, to preserve the feasibility of the high-level footstep plan, we require that each body and end-effector movement segment start and end at the state determined by the footstep planner. This is accomplished by constraining $S_\text{start}$ and $S_\text{goal}$. 
An example trajectory of body movement is shown in Fig.~\ref{fig:motion_planning}(C). 

We employ a similar technique for free end-effector movement by warm-starting SCP with a locally sampled collision-free straight-line gripper trajectory between anchor points. The optimization is nearly identical to that for body movement, but with an external wrench term imparted by the cantilevered boom and end-effector included in the dynamics. An example end-effector trajectory is shown in Fig.~\ref{fig:motion_planning}(D).

ReachBot's locomotion is contingent on reliable grasps, so a robust motion plan must incorporate knowledge of grasp strength and variability. 
It is therefore essential to have an accurate model to predict the force limit surface of each anchor point. Additionally, we seek to expand ReachBot's accessible terrain by designing a gripper that can withstand large pulling forces and conform to many different rock geometries with a lightweight design. Finally, our planner relies on knowledge of anchor point locations beforehand (Fig.~\ref{fig:motion_planning}(A)), requiring a perception system to aid in grasp site identification. In the following sections, we present solutions to each of these challenges to support ReachBot's locomotion in realistic environments.

\subsection*{Gripper modeling and grasp prediction through limit surfaces } 
\label{subsec:limitsurface}

To plan a trajectory through a rocky cave, one must estimate the reliability of possible grasp points. As part of this process we need to estimate the maximum pull force that a boom and gripper can exert while grasping a rock feature. As noted in previous work \cite{asbeck2012designing,jiang2018stochastic,iacoponi2020simulation}, grasping rock surfaces with microspines is inherently stochastic but can be modeled with a defined mean and standard deviation. \rev{In contrast to some previous applications involving arrays of microspines \cite{asbeck2012designing, wang2017design, parness2017lemur, RuotoloRoigEtAl2019}, we do not assume that the surface is nearly flat at the length scale of the gripper; instead, we seek rounded features that the gripper can partially enclose. This condition increases the pull force for a given number of spines.} 



\begin{figure}[H]
    \centering
    \includegraphics[width=0.95\textwidth]{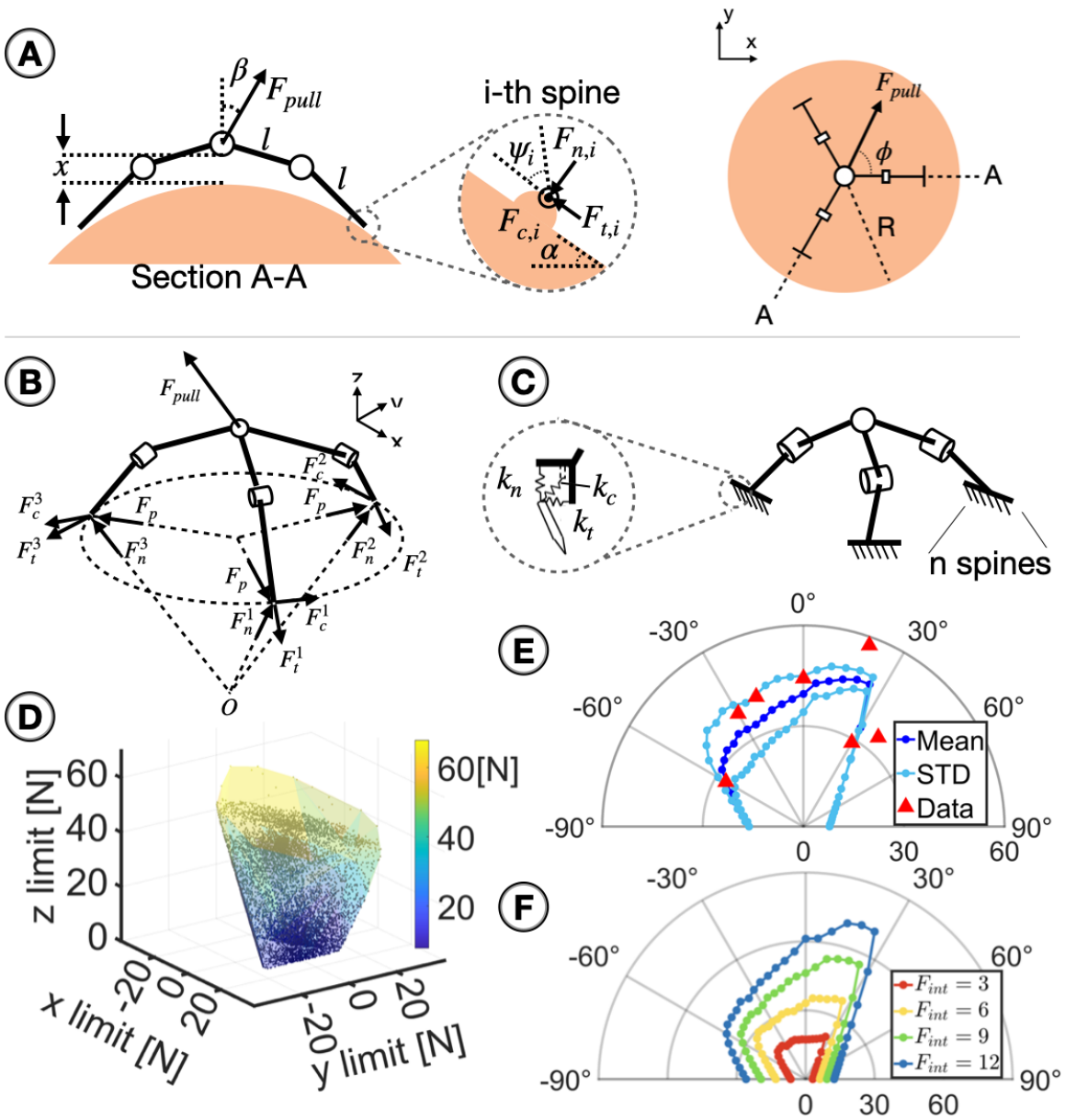} 
    \caption{\textbf{Grasp model and analysis.} (A) Geometric parameters for three-finger grasp model.(B) Free body diagram and contact forces. (C) Compliant spine-asperity contact.  (D) A 3D limit surface generated from a Monte Carlo simulation with 10,000 samples using parameters derived during the field test; \rev{force components are with respect to ($x,y,z$) coordinates in (B), in a frame centered at the wrist.}
    (E) Cross-section polar plot of the magnitude of applied pull force, $F_{pull}$, using parameters from field test, with $(-90^\circ \leq \beta \leq 90^\circ)$, where $\beta$ is the pull angle, and assuming 20 spines engaged \rev{(25\%)}. Slight asymmetry in $\alpha$ arises from asymmetry in the rock. Straight sides correspond to failure associated with a finger lifting off. Red triangles mark field test results. Additional details are provided in Supplement \ref{sup:MC}. (F) Cross-section polar plot showing the dependence of $F_{pull}$ on grasp force, $F_{int}$.
    }
    \label{fig:3d_eq}
\end{figure}

To estimate the pull force limit distribution, we perform an analysis wherein we assume the gripper is grasping a rounded rock feature which can be approximated as a spherical patch with radius $r$. The gripper is symmetric, with three fingers, each having two links of length $\ell$ (Fig.~\ref{fig:3d_eq}). For a locally spherical surface, links of equal length provide the tightest fit to the surface. A pulling force, $F_{pull}$, is applied to the wrist joint with a direction defined by the two angles $\beta$ and $\phi$. When the gripper is loaded, the wrist joint will be a small distance, $x$, away from the surface.

We now consider a single spine that is part of a small array of spines at one of the fingertips. As in 
\cite{asbeck2012designing,jiang2018stochastic,iacoponi2020simulation}, the spine tips do not penetrate the rock surface and are subject to Coulomb friction. Thus, for each spine, the ability to sustain loads $F_{ti}$ and $F_{ci}$ in the local tangential directions is a function of the normal force, $F_{ni}$, and the angle of inclination at which it makes contact. 
The overall angle is a function of where the spine rests on an asperity (illustrated as a small hemispherical bump in Fig.~\ref{fig:3d_eq}(A)), parameterized by 
$\psi_i$, and by the local surface normal in the vicinity of the asperity, which is parameterized by $\alpha_i$. The angle $\alpha_i$ is a function of how much the gripper encloses the rocky feature. For a flat surface $\alpha=0$; if $r<\ell$, $\alpha \rightarrow \pi/4$. The normal force, $F_{ni}$ is a function of both the applied load to the fingertip and the internal grasp force, $F_{pi}$, which we can control. Computing the force balance and assuming Coulomb friction, we have the no-slip condition for a spine:
\begin{equation}
        \frac{\sqrt{(F_{t,i}\cos\psi_i - F_{n,i}\sin\psi_i + F_{p,i}\cos(\alpha+\psi_i))^2 + (F_{c,i})^2}}{F_{t,i}\sin\psi_i + F_{n,i}\cos\psi_i + F_{p,i}\sin(\alpha+\psi_i)} \leq \mu.
        \label{eq:non-slip_3d}
\end{equation}
\rev{The derivation of this expression can be found in Supplementary Methods. We note that the no-slip condition depends not only on the asperity distribution but also on $\alpha$, which increases as the surface becomes increasingly convex. This dependency, along with the grasp force balance equations described below, accounts for the ability to provide larger pull-off forces for a given grasp force when grasping convex features. A similar relationship has been noted for perching air vehicles that use claws to grasp branches or tree trunks \cite{roderick2017bioinspired}.}

For an array of spines, there will be a distribution of asperities, leading to a distribution of angles with corresponding variations in the maximum tangential forces. If we assume a
distribution of angles, $\psi_i$, we can compute the overall contact force limits for an array of spines at each fingertip, following a procedure similar to that in 
\cite{wang2019spinyhand,jiang2018stochastic,iacoponi2020simulation}.

Whether the array of spines at each fingertip slips or holds also depends on the overall force balance for the hand. The hand has a ball joint at the wrist, so \rev{no torques are} transmitted. 
We construct a grasp matrix, assuming that the spine array at each fingertip is compact enough to treat it kinematically as a point contact. This is further justified by how the spines are supported in compliant suspensions so the spine array does not produce substantial bending moments about the contact. Given an assumed external force $F_{pull}$ and known geometry, we have nine quantities to solve for, corresponding to the normal and tangential force components at each contact. Force equilibrium provides six equations. In addition, we have three kinematic equations stemming from the condition that spines do not penetrate and should not lift off from the rock surface. This rigid body condition can be converted into equations on the forces using the known suspension stiffnesses ($k_n, k_t, k_c$) at each spine. The result is an invertible system:
\begin{equation}
    \mathbf{A} F = B,
    \label{eq:3d_symm}
\end{equation}
\noindent where $\mathbf{A}$ is a 9x9 compliant grasp matrix that maps contact forces to resultant forces at the wrist, $F$ is the vector of contact forces to be solved, and $B$ is a 9-element vector containing terms with $F_{pull}$ and the pull angles, $\beta$, and $\phi$. 
Details of the derivation are provided in Supplementary Methods.

Eq. (\ref{eq:non-slip_3d}) and (\ref{eq:3d_symm}) can be computed for a given grasp geometry and gripper, summing the contributions for arrays of spines and assuming a distribution of asperities, parameterized by $\psi_i$. We perform a Monte Carlo simulation by running this computation over many samples of asperity angles to approximate the distribution of the maximum magnitude of the pulling force, then by sweeping over the direction of the pulling force ($\beta$, $\phi$) to generate a grasp limit surface. Details of the Monte Carlo simulation are provided in the Supplementary Methods. 
A sample limit surface is shown in Fig.~\ref{fig:3d_eq} using the following parameter values: $n_{spines} = 20$ , $\alpha_1 = 1.22~\text{rad}$, $\alpha_2 = 0.52~\text{rad}$, $\alpha_3 = 0.70~\text{rad}$, $\mu = 0.39$, $r = \SI{11.6}{\centi\meter}$, $x = \SI{4.5}{\centi\meter}$, $F_{int} = \SI{6}{\newton}$, $F_{max} = \SI{12}{\newton}$ (maximum tangential force for a single spine to prevent permanent bending), $k_n = \SI{15}{\newton\cdot\meter^{-1}}$, $k_t = \SI{3e6}{\newton\cdot\meter^{-1}}$, $k_c = \SI{1}{\newton\cdot\meter^{-1}}$, $n_{iterations} = 200$. \rev{These values correspond to our estimates for the field tests on lava rock using data from a lava tube, obtained by the perception system as described in the section on grasping site identification.}

The limit surface model is applied during motion planning to evaluate potential grasp sites based on rock geometry and expected pulling directions. The model is also useful to compare limit surfaces of different gripper designs for a given set of grasp geometries, thus aiding in selecting a gripper design for a specific application.

\subsection*{Gripper design}
\label{sec:gripperdesign}

The primary design requirements for the grippers include: low weight, high conformability to rock features, and large \rev{direction-dependent maximum pullout force as quantified by the limit surface. The evaluated force directions should correspond to the anticipated range of boom pulling directions.} The need to minimize weight is driven by two considerations: a heavier gripper produces higher bending loads that may cause the booms to buckle, and increases the torque required by the shoulder actuators to steer the boom. 

\begin{figure}[h!]
    \centering
    \includegraphics[width=1\textwidth]{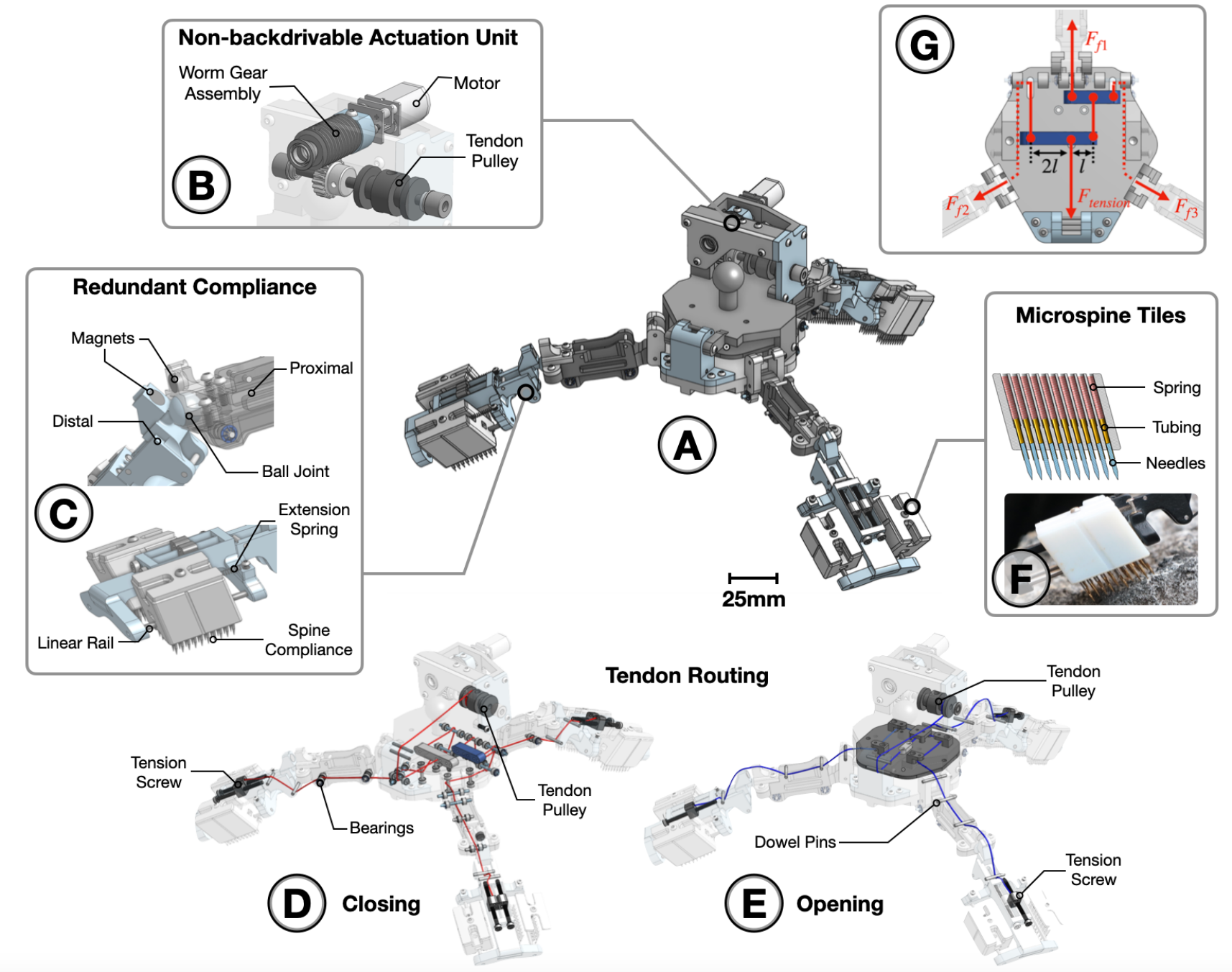}
    \caption{\textbf{Mechanical gripper design.} (A) Opening and closing tendons are driven by a single gearmotor and drum (B). Proximal/distal phalanges are connected by ball joints, with permanent magnets to hold them in place when the hand is open (C). The opening and closing tendons respectively are illustrated in (D) and (E). Linearly constrained spines (F) are arranged in compact arrays on carriages that float with respect to the fingertips. Additional details are provided in Supplementary Methods.}
    \label{fig:gripper}
\end{figure}

To meet the requirements, we designed and tested an underactuated, tendon-driven gripper with microspines \rev{(Fig.~\ref{fig:gripper}(A)).} The gripper has three fingers, each with two phalanges. To save weight, the entire gripper is actuated by a single motor \rev{(Fig.~\ref{fig:gripper}(B))}.  As in some other underactuated grippers \cite{catalano2014adaptive, RuotoloRoigEtAl2019}, the main tensile loads are borne by the tendons and not the joints. 




Two sets of tendons are used to close and open the fingers (Fig.~\ref{fig:gripper}(D,E)). The tendons are driven by a motor with a non-backdrivable transmission and high gear ratio, which permits moderate grasp forces
($\approx 7.5$\,N on each finger), albeit with slow operation ($\approx$ 20\,s to close the hand from a fully opened configuration). The closing and opening tendons are wrapped around a drum that is fixed to a shaft, with a worm gear assembly (Fig.~\ref{fig:gripper}(B)). The two primary tendons are wrapped around the drum separately, one clockwise and one anticlockwise \rev{for opening and closing, respectively}.
As in many underactuated hands, a differential mechanism divides the tendon forces among the fingers for closing \cite{baril2010static}. In the present case we use a three-way whiffletree, illustrated in Fig.~\ref{fig:gripper}(G). 

A known issue with compliant underactuated hands is that one wants the proximal joints to start closing before the distal joints in order to achieve a large ``acquisition region'' \cite{baril2010static,belzile2014stiffness}. In hands that use torsion springs or other elastic elements, this requirement can lead to either floppy proximal joints in the relaxed state or stiff distal joints, which increases the effort required for grasping. Solutions can include exploiting nonlinearities in the joint stiffness  \cite{stuart2017ocean}. 
Here, we use pairs of permanent magnets at the medial joints (Fig.~\ref{fig:gripper}(C)) that hold the fingers precisely in the fully extended position until tension in the closing tendon \rev{causes the joint torque to exceed a threshold determined by the strength of the magnetic force---by which time the proximal joints have already started to flex. With this provision, the stiffnesses of the joints can be adjusted to
prevent the distal phalanges from curling inward prematurely when grasping small convex features. Similar uses of magnets to achieve a nonlinear stiffness in underactuated hands can be found in \cite{gerez2019employing,gafer2020quad}.}

\begin{figure}[h!]
    \centering
    \includegraphics[width=0.9\textwidth]{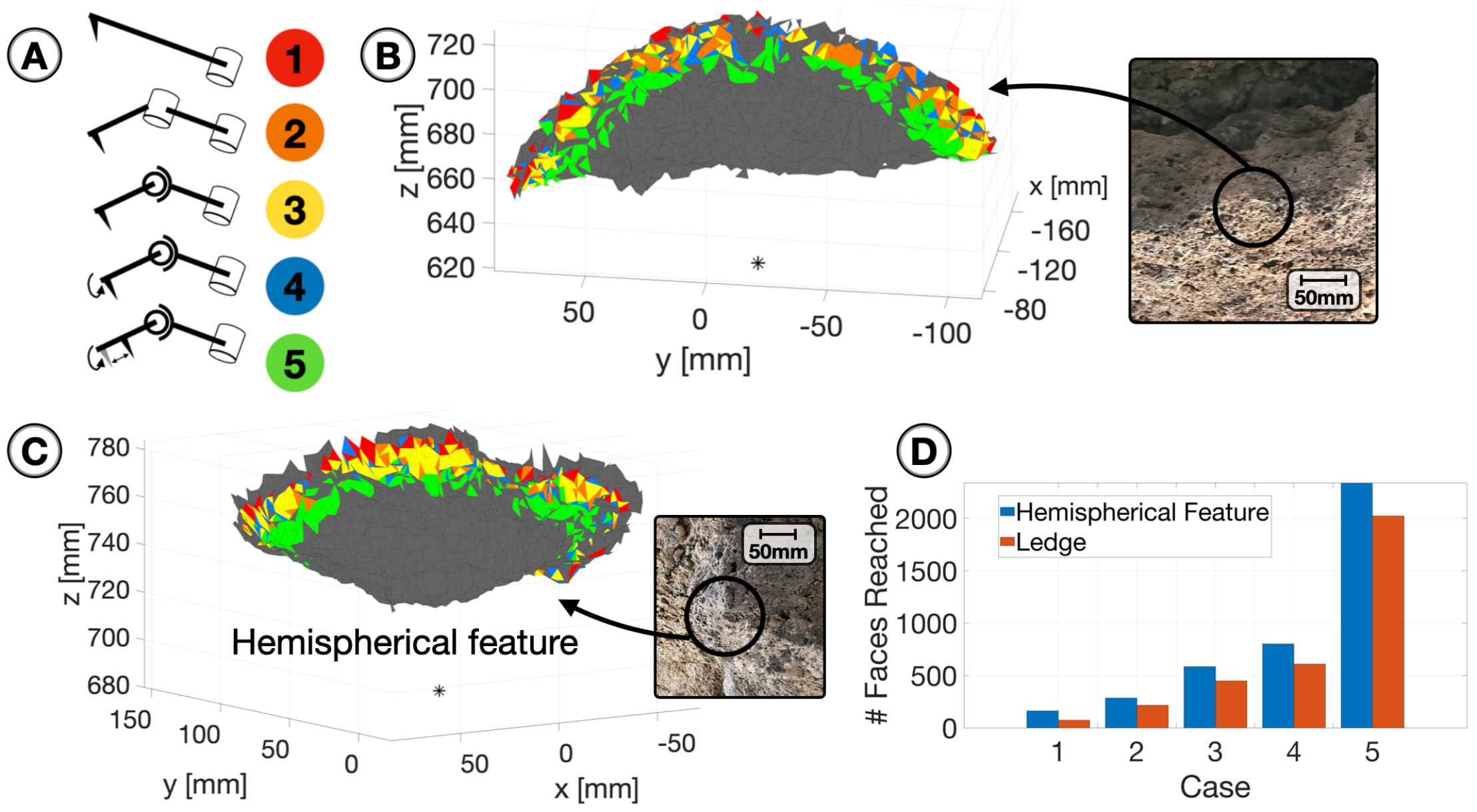}
    \caption{\textbf{Reachability analysis of different gripper configurations.} (A) on representative lava rocks presenting hemispherical (B) and ledge (C) surface patches scanned and meshed at $\approx 0.1$\,mm resolution (insets shows actual surface).
    Red: In the first case, there is one phalange with the spine fixed at the end. Orange: In the second case, there are two phalanges connected with a revolute joint. Yellow: In the third case, there are two phalanges connected with a ball joint. Blue: The fourth case is the same as the third case with the spine able to rotate with respect to the distal phalange. Green: The fifth case is the same as the fourth case with the spine having up to 10\,mm tangential and normal linear travel with respect to distal phalange.
    On meshed surfaces, each colored region depicts the additional graspable area with respect to the previous case. (D) Comparison of reachable areas for the different cases for each surface.
    }
    \label{fig:reachability}
\end{figure}

Rock surfaces, especially lava rocks, are highly irregular both at macroscopic and microscopic scales. Therefore, compliance is built into the hand at multiple length scales. \rev{At the microscopic scale 
($<$ \SI{100}{\micro\meter}), asperity size determines the required spine tip radius. In this regard, the rock samples in a lava tube are similar to other rocks climbed by robots with spines ~\cite{asbeck2006scaling,parness2017lemur,RuotoloRoigEtAl2019} and the spine tip radius ($\approx$ \SI{10}{\micro\meter}) is comparable. At a slightly larger scale, surface waviness determines how much linear travel the spines should have to ensure that enough spines on each tile will contact the surface. We assume $\approx 25$\% asperity engagement in the analysis in the previous section. A surface scan of the rock in Supplementary Methods 
reveals that there is substantial waviness at the approximately \SI{1}{\centi\meter} scale. To accommodate this waviness the spines have 10\,mm linear travel. In practice, we have observed empirically that at least 25\% of the spines do actively engage. The third relevant scale of surface waviness is at 
length scale of the hand. We rely on the existence of rounded features and we design the hand to be able to confirm to these features so that all the fingertips make contact. This topic is addressed further in the reachability analysis in the next section and in Fig.~\ref{fig:reachability}.
} 

At the tip of each finger are a pair of compliantly supporting microspine arrays (Fig.~\ref{fig:gripper}(C)).
As described in \cite{wang2017design,wang2019spinyhand}, arrays of linearly constrained spines provide a high packing efficiency for maximum shear force per unit area.  To provide more compliance and conformability than in these earlier designs, the arrays slide semi-independently on rods, with a compliant coupling. \rev{Each floating carriage is connected to the distal structure through one rod, thus allowing for both linear compliance and torsional compliance with respect to the finger.} These suspension elements, along with a small amount of compliance for each spine \rev{(Fig.~\ref{fig:gripper}(F))}, provide load sharing so that the first few spines to make contact with asperities do not take most of the load and fail prematurely. \rev{A similar floating tile design is seen in a climbing robot \cite{zi2023mechanical}.}

As noted in the previous section, we exploit rounded rock features to increase the ratio of pull-off force to grasp force. Features can exhibit approximately spherical or cylindrical geometries, for knobs or ledges, at the length scale of the hand. To accommodate both of these geometries the fingers have ball joints that allow bending and twisting. With this flexibility, \rev{the gripper can passively assume either an 
approximately symmetric three-fingered grasp or an opposed grasp, with two fingers opposite the third finger, as it is closed and loaded.} 

A question not yet addressed is how many degrees of freedom the gripper fingers require and how many phalanges they should have. To address this question we conducted numerical studies with typical lava rocks for a single finger, as illustrated in Fig.~\ref{fig:reachability}. For simplicity, there is only one spine at the end of the finger---\rev{we assume there is enough compliance in the spine tiles and spines that we can obtain roughly 25\% spine contact for the range locations and angles simulated for a single spine.}
Even so, the search space is large and an exhaustive search is prohibitive. However, for one-phalange and two-phalange designs with different joint types, we can examine the ability to conform to rock surfaces with typical (partial) spherical and cylindrical shapes. For effective grasping, each spine will need to meet the surface at a range of angles between \SI{10}{\degree} and \SI{30}{\degree} from the surface normal.  Fig.~\ref{fig:reachability} illustrates five cases \rev{with sequentially increasing mobility}. In the first case, there is one rigid phalange with a revolute joint at the base. In the second case, two phalanges have a revolute joint between them. In the third case, there are two phalanges with a spherical joint in between. The fourth case is the same as the third case with an additional rotation degree-of-freedom at the spine. The fifth case is the same as the fourth case with additional shear compliance of the spine. Each shaded region highlights the additional surface area that can be grasped in comparison to the previous case. As illustrated in Fig.~\ref{fig:reachability}(D), the fifth case represents a substantial improvement over the previous cases, which is why it is incorporated into the gripper design.

The prototype gripper presented in this paper reflects the design requirements for grasping rough convex rock surfaces with high reliability. Force tests were limited to avoid breaking the 3D printed plastic parts. We expect the trend of increasing $F_{pull}$ with $F_{int}$ seen in Fig.~\ref{fig:3d_eq}(F) to \rev{scale to larger loads, limited by the chosen motor, tendons, finger components, and surface asperity strength.}

\subsection*{Grasping site identification and selection}
Grasp strength is strongly dependent on surface profile, but the repositioning, detaching and deployment of a gripper toward a new site can be a time-intensive maneuver \rev{limited mainly by the speed of boom steering and deployment}. 
Therefore, it is beneficial for ReachBot to have high confidence in a potential grasp before committing to extending its end-effector.
Our solution is a two-stage \rev{image-based} surface scanning process: a cursory scan from the body provides rough indications of suitable grasp sites, such as convex features. It then samples from an extended boom refines the grasp site evaluation. In this strategy, there are two main geometric parameters that must be determined for the subsequent limit surface analysis.

\begin{figure}[h!]
    \centering
    \includegraphics[width=0.95\textwidth]{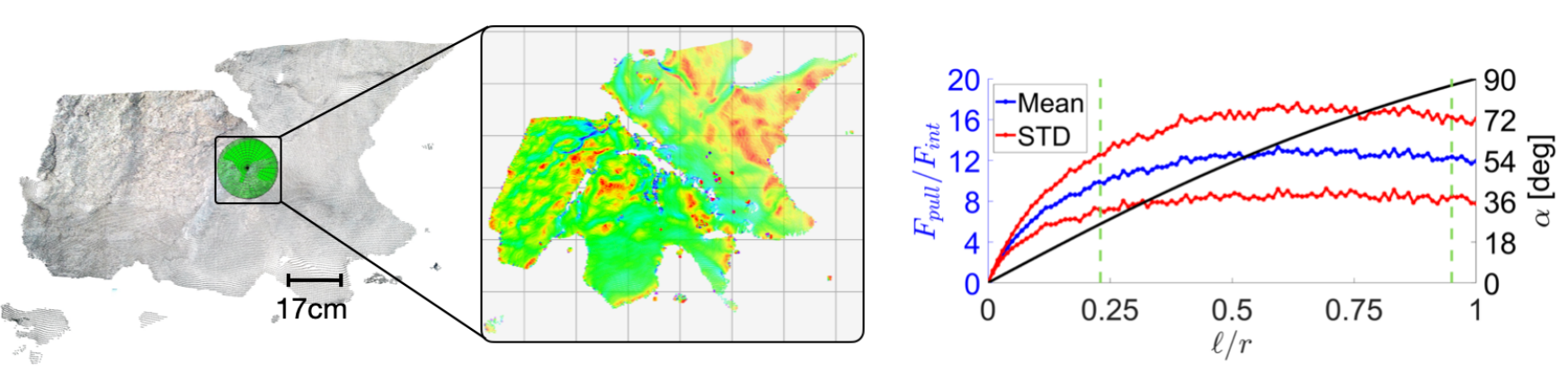}
    \caption{\textbf{Two-stage perception and detection of suitable grasp sites.} (A) From a stowed position away from the rock wall, raw camera depth data is processed to identify a promising region, with roughly spherical features. Inset shows an image of the deployer in a stowed position collecting data from a rock wall. (B) The boom is deployed for a closer scan, from which a local $\alpha$ angle heat map is generated. Red surfaces are nearly flat at the scale of the gripper; green regions correspond to a range of values $\alpha_{min} < \alpha < \alpha_{max}$, where $\alpha$ is the local surface normal in the vicinity of the asperity, that are likely to yield strong grasps. (C) Shows a plot for $\beta=0$, where $\beta$ is the pull-off angle, of the computed ratio of pull-off force to grasp force for different $\ell/r$ values, where $\ell/r$ is the ratio of link length to rock radius. The desired range of $\ell/r$ values are marked (related to $\alpha$ values). 
    }
    \label{fig:perception}
\end{figure}

The first objective is to find suitable convex surfaces. For the present, we approximate these by fitting spheres of radius $r$ onto protruding rocks from a distance. Although there are many methods for spherical approximation, we use the M-estimator SAmple Consensus (MSAC) algorithm for sphere fitting, a variant of RANdom SAmple Consensus (RANSAC). \rev{RANSAC is a widely used method for fitting geometric shapes onto data \cite{fischler1981random,schnabel2007efficient}, and we chose the MSAC variant for better performance with noise \cite{torr1997robust,torr2000mlesac}. We constrain the output with a window of desired values for $\ell/r$, which relates to $\alpha$ as shown in Fig.~\ref{fig:perception}(C), with the computation using the same parameter values as the limit surfaces plotted in Fig.~\ref{fig:3d_eq}. 
This plot also shows that the ratio of pull-off force to grasp force grows rapidly as $\alpha$ and $l/r$ increase from the case of a flat surface ($\alpha=0$) to a rounded feature and plateau well before the local surface normals at the fingertips become radially opposed.
}
Details of the relationship between $\ell/r$ and $\alpha$ are provided in Supplementary Methods.
Because ReachBot has a large workspace, we assume that we can afford to be selective when determining grasp sites. If multiple nearby grasp sites are available, they are ranked based on computed limit surface models and the \rev{preferred loading direction(s), which will depend on the stance and desired trajectory of the robot.
For field tests reported in the next section, there is a single loading direction corresponding to the fixed angle of the boom.}

Once a boom is close to a region of interest, a second objective is to determine more precisely the value of $\alpha$ to compute the limit surface and allow for readjustment before committing to a grasp. The $\alpha$ angles are calculated as the angle between the gripper reference axis and the estimated normal vector for the points in the point cloud. ReachBot commits to a grasp if $\alpha$ falls within a range of acceptable values between $\alpha_{min} = 25^\circ$ \rev{(below which the pull force drops rapidly for most pulling angles)} and $\alpha_{max} = 85^\circ$ \rev{(above which there is not substantial improvement in pull force)}. \rev{These bounds are chosen based on when the ratio of pull-off force to grasp force plateaus (Fig.~\ref{fig:perception}(C)).} The upper bound is further dictated by the camera's inability to discern a steep ledge from a potentially discontinuous surface. Mapping $\alpha$ angles spatially can also help categorize the convex surface as roughly hemispherical or as a ledge, which informs the limit surface and subsequent motion planning.

Fig.~\ref{fig:perception} illustrates the main perception pipeline principles used to detect a grasp site. First, from a distance, potential anchoring sites are determined by approximating them as convex surfaces with a radius $r$, with only promising $\ell/r$ ratios kept. Fig.~\ref{fig:perception}(A) highlights a convex surface with estimated radius $r = \SI{17}{\centi\meter}$. Once the boom has extended towards an anchoring site, more detailed depth data is gathered to determine the angle $\alpha$ for the limit surface model, as shown in Fig.~\ref{fig:perception}(B). The calculated $\alpha$ angles for a successful grasp on this rock were between $55^{\circ}$ and $60^{\circ}$ for each microspine tile, suggesting that the site is indeed a gently curved hemispherical surface.  The two-step perception strategy was also used to compute the $r$ and $\alpha$ values to generate the limit surface plots in Fig.~\ref{fig:3d_eq}(D-F). The components of the perception system illustrated in Fig.~\ref{fig:perception} correspond to the field test site. 

\subsection*{Deployment in a lava tube}

As noted earlier, planetary caves and lava tubes are among the most promising geological and astrobiological targets in the solar system \cite{blank2018planetary,titus2021roadmap,wynne2022planetary,wynne2022fundamental}. Given that martian lava tubes are of particular interest \cite{LeveilleDatta2010, phillips2020mars} it was decided that a field test of ReachBot technology should be conducted in a reasonable analog on Earth. To this end, tests were conducted in an unnamed lava tube near Pisgah Crater, within the Lavic Lake volcanic field in the Mojave Desert, California (\SI{34.74977}{\degree},\SI{-116.36733}{\degree}). 

The Lavic Lake volcanic field consists of Quaternary (likely late Pleistocene \cite{phillips2003cosmogenic}) basaltic pahoehoe and aa lava flows \cite{wise1966geologic} and has a long history of serving as a planetary analog \cite{greeley1988relationship,arvidson1998rocky}. The presence of both pahoehoe and aa flow textures ensures that the ReachBot prototype is subjected to a wide range of surface roughnesses, similar to what could be encountered on Mars. The specific lava tube where field testing was conducted was chosen after an initial survey of local lava tubes based on practical considerations such as entrance size, ease of access, and safety to human operators.

The partial ReachBot prototype in the field test consists of a single boom deployer mounted on a shoulder joint that allowed it to pan and tilt. The gripper, presented in Fig.~\ref{fig:gripper},
is mounted at the end of the boom along with an Intel Realsense d455 camera. The entire assembly is mounted on a tripod, representing the body of ReachBot\rev{, and is controlled via teleoperation}. The overall assembly is shown in Fig.~\ref{fig:glamour}(B). A fully functional ReachBot will have eight of these shoulder assemblies supporting a central body. 

\begin{figure}[h!]
    \centering
    \includegraphics[width=0.7\textwidth]{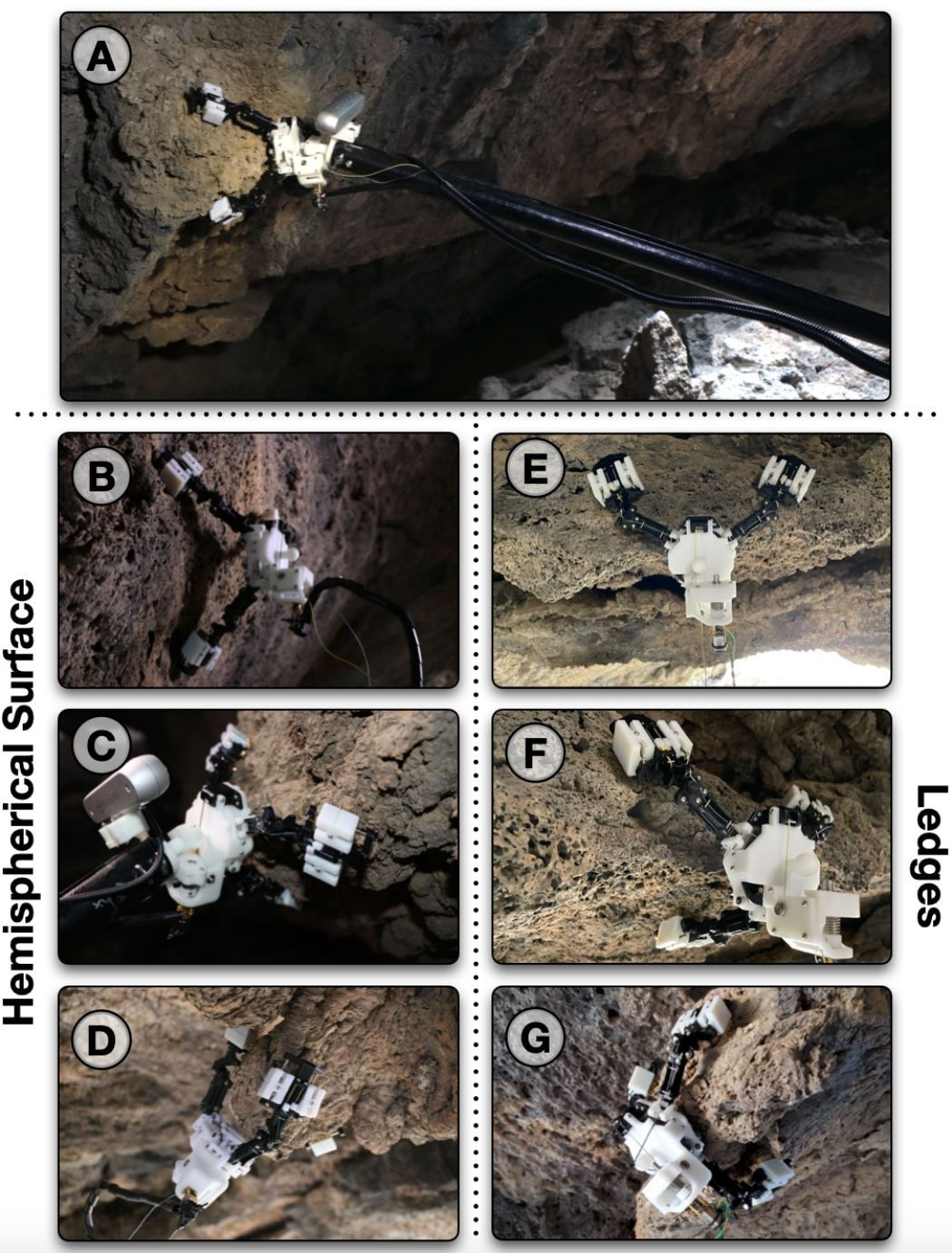}
    \caption{\textbf{Field test demonstrations at a lava tube in the Mojave Desert.} (A) gripper deployed at end of boom after identifying a target. (B-D) Locally spherical rock features (E-G) locally cylindrical features. In images (B-G) boom is removed for pull tests with force gauge.}
    \label{fig:fieldtest}
\end{figure}

The single-arm prototype was able to target, deploy and grasp many target sites within the lava tube. One full sequence is shown in the supplementary video.
For the perception system, the field test confirmed the advantages of a two-step perception strategy, starting with a remote scan \rev{with the RGBD camera} (Fig.~\ref{fig:perception}(A)). \rev{The remote scan, conducted under teleoperation, identified areas that were expected to yield} fruitful grasp locations \rev{(Fig.~\ref{fig:perception}(B))}. Figure \ref{fig:fieldtest} panels (B-D) and (E-G) show grasps on features with approximately spherical (radially symmetric) and cylindrical geometry, respectively. In these images, the boom has been removed for measuring pull-off forces with a Mark-10 (model M4-50 with $\SI{0.1}{\newton}$ resolution) force gauge. \rev{The gripper is loaded in the direction of the boom.}
In all cases, the grasps achieved at least $\SI{34.2}{\newton}$ (larger forces were not applied to avoid breaking the 3D-printed plastic components of the prototype during field testing). 

Additional pull-off tests using only the gripper were subsequently applied by hand using a large rock sample acquired at the testing site. In these tests, the gripper was placed on a surface of the rock and then pulled until the grasp failed, either from part of the 3D-printed structure failing or from the spines breaking or slipping. The maximum force in these tests was recorded, again using the Mark-10 gauge and the pulling angle was recorded from digital photographs. These results are plotted on the modeled limit surface for the same geometry and gripping parameters in the limit surface in Fig.~\ref{fig:3d_eq}(E).


\section*{Discussion}


We have presented a robot capable of navigating scientifically promising but previously inaccessible areas, such as planetary caves and lava tubes. The work in this paper extends preliminary studies using a simplified planar ReachBot model \cite{SchneiderBylardEtAl2022, ChenMillerEtAl2022} to a full 3D analysis with a realistic consideration of grasp strength and grasp site selection. The results presented here also include field testing because the details of sensing and grasping rock surfaces characteristic of lava tubes are difficult to address purely in simulation. In comparison, many of the motion planning and control considerations can be addressed in simulation once accurate grasp models are available.

Based on considerations of grasping the rocks in lava tubes, we adopted a strategy of seeking convex features that afford a high ratio of pull-off force to grasp force. To this end, we developed a model of a theoretical three-fingered microspine gripper which can be used to estimate a force limit surface based on different surface geometries and gripper parameters. Based on these principles, we prototyped a highly compliant under-actuated microspine gripper. For field tests we also developed a two-step perception system that identifies possible grasp locations, trading density of targets for high ratio of pull-off to grasp force. Finally, we tested a single-arm prototype in a lava tube in the Mojave Desert. During this field test, the ReachBot prototype identified, targeted and grasped many sites, demonstrating the potential capability of a full ReachBot system while also identifying some real-world challenges for future mission success.

There remain many challenges and areas of future work in motion planning, mechanism design, and perception.
First, the dynamic model should include the distributed mass of the booms. Additionally, the control effort metric $c(\tau_k)$ currently penalizes prismatic force and motor torque equally, but should balance the relative risk of failure from high prismatic forces (which could cause gripper failure) with the failure from high motor torque (which could cause the boom to buckle). 

Second, due to the stochastic nature of spine-surface interactions, grasp failure is an eventuality that must be considered in planning. A robust motion plan must both ensure that the probability of failure is low and that if a grasp does fail, a contingency trajectory exists to stabilize ReachBot with low likelihood of additional failures. We have identified these challenges in a simplified planar model \cite{NewdickOngoleEtAl2023} but need to extend them to 3D. Numerous improvements to the computational efficiency of the grasp force computation and motion planning are also possible.

Third, to support the force control assumed by ReachBot's  motion plan, each boom will need to have force sensing capability. Force sensing will also help to avoid overloading components. In addition, a stronger and more reliable boom deployer will be needed for future prototypes. \rev{Based on different mission specific gripper requirements, stronger grippers with metal parts, stronger (or more numerous) spines, and a larger motor can be chosen.} A finding from field tests is that it may be desirable to add an actuated roll degree of freedom to better align the gripper with asymmetric features. This addition will increase the distal inertia but is probably worthwhile to increase the density of grasp targets.

\rev{Fourth}, because ReachBot is a robot with exceptional workspace, its perception system must be designed for its unique form factor. In addition, although the core principles of two-stage sensing were verified, a more flexible shape approximation approach will be investigated. Forthcoming work includes a wider range of geometric feature descriptors, targeted illumination for sensing at different ranges, and inquiry into other long-range sensing beyond the Realsense camera. We will also investigate deep learning approaches for grasp site identification.

\rev{Further, a hybrid boom-and-cable design is an interesting possible direction for ReachBot. For example, booms can be used to position and attach anchoring grippers with tensioned cables. Accordingly, one or two booms per robot simplifies the articulation of a full system while maintaining a large workspace. Another advantage is that cables could cover even greater distances than extendable booms. A hybrid ReachBot could use a boom to position a gripper while the robot is closer to a desired surface, and pay out cables that allow the robot to move much further away than the length of the original boom.}

\rev{Finally, we note that ReachBot's characteristic parameters can be scaled based on the expected environment. For example, boom size relates to gripper size, which relates to tile size and graspable feature size. The scaling chosen for this field test is a good match for the surface characteristics of lava tubes in the Mojave Desert, which are discussed in Supplementary Methods. 
Accordingly, we may change the entire scale of ReachBot for operation on Mars, the moon, or other extraterrestrial bodies.}

In summary, many extensions are possible. Nevertheless, the work presented in this article provides the basis for the motion planning of a small robot with long arms terminating in grippers that grasp on rocky terrain. Results from the field test confirm the predictions of maximum grasp forces and underscore the importance of identifying and steering towards convex rock features that provide a strong grip. They also highlight a characteristic of grasp planning with ReachBot, which is that identifying, aiming for, and extending booms involves a higher level of commitment than grasping objects in manufacturing scenarios. 



\section*{Materials and Methods}

\paragraph*{3D simulation environment}
To support our need for a simulation environment with variable-length links and hybrid dynamics, we built a custom physics-based simulation in Python. We used the Rospy package to support visualization using RViz as well as future hardware integration. We implemented sequential convex programming optimization using CVXPY, and used standard propertional-integral-derivative control for trajectory tracking.
The simulation environment was generated to match our field test site, and constants such as pan and tilt actuator limits were chosen to match our real boom. Parameters used in the simulation are shown in Table~\ref{table:sim_constants}.


\paragraph*{Spine array fabrication}
Each microspine used was manufactured from a leather sewing needle with a titanium-nitride finish (Organ Needle HAx130N PD). The sewing needle tip was cut from the stem at \SI{7}{\milli\meter} length, then inserted into a stainless steel tube (McMaster Carr 8987K521), secured in place with superglue (Loctite 401). The stainless steel tube was then inserted into a compression spring (Lee Springs CIM010ZL 05S) with a \SI{3}{\milli\meter} overlap, also secured with superglue (Loctite 401). The spine suspension assembly then was placed into a 3D-printed (Stratasys Objet VeroWhite) spine tile array where the inside lip of the hole holds the edge of the stainless steel tube in place (Fig.~~\ref{fig:gripper}(F)).    

\begin{table*}[h]
\begin{center}
\begin{tabular}{ |l|c c c c c c c| } 
 \hline
 Constants & $G$ & body length & body diameter & $\text{m}_\text{body}$ & $\text{m}_\text{gripper}$ & & \\
 & $3.721$m/s$^2$ & $0.8$m & $0.4$m & $10$kg & $1$kg & & \\ 
 \hline
 Joint & $\theta_\text{min}, \theta_\text{max}$ & $\phi_\text{min}, \phi_\text{max}$ & $b_\text{min}$ & $b_\text{max}$ & $F_\text{min}$ & $F_\text{max}$ & $M_\text{min}, M_\text{max}$\\ 
 Limits & $\pm 180^\circ$ & $+ 90^\circ$ & $0.2$m & $10$m & $0$N & $40$N & $\pm 10$Nm\\ 
 \hline
\end{tabular}
 \caption{Parameters used in ReachBot simulation.
}
  \label{table:sim_constants}
\end{center}
\end{table*}

\paragraph*{Gripper fabrication}
Most of the gripper components were manufactured through 3D printing, using three different printers. The spine tiles were printed with Stratasys Objet VeroWhite for the dimensional accuracy required. The fingers were printed with HP Multi Jet Fusion (MJF) Nylon 12 PA. The rest of the gripper was printed on a Formlab3+ with Rigid 4k material for rigidity. The connections between parts were accomplished by press-fitting brass screw-to-expand inserts. The motor (Pololu 1000:1 Micro Metal Gearmotor) underwent a further 30:1 gear reduction through a wormgear assembly (Misumi SUW0.5-R1 and G50B20+R1) and controlled the loading tendon (Twinline braided Vectran 125) and opening tendon (Power Pro Spectra 40\,lbs (178\,N)).

\paragraph*{Limit surface experimental setup}
To validate the limit surface,  pull tests were performed with the gripper on volcanic rocks collected from the field test site. A digital force gauge (Mark-10, model M4-50, $\SI{0.1}{\newton}$ resolution) was connected to the gripper along a tendon. The angle of the force applied was measured by analyzing photos taken during the experiment. We note that tests to failure typically blunt the spine tips, unless something else fails first. Therefore the probability of each spine finding an asperity gradually diminished as these tests continued. When operating the robot one would seek to avoid such failures.

\paragraph*{Grasp reachability analysis}

To analyze reachbability, we obtained a triangulated mesh of a scanned rock surface, then choose a point slightly above the rock surface to position the base of a simulated finger, then examined five cases as described in Fig.~\ref{fig:reachability} using Algorithm \ref{alg:reachability}.
We sampled each additional degree of freedom at 5 degree increments and translation in \SI{2}{\milli\meter} increments.

We then followed Algorithm \ref{alg:reachability} to look for rock faces that are reachable by the spine, at which the spine attacked the face at the desired angle (between \SI{10}{\degree} and \SI{30}{\degree} from the surface normal) and the finger itself did not collide with the rock.


\paragraph*{Perception system field test}
The perception field test used an Intel Realsense d455 RGBD camera which was rigidly mounted to the distal end of the boom, co-located with the gripper wrist. The camera provided data from a range of $0.6$ - $\SI{6}{\meter}$. 

The point cloud was down-sampled using a voxel size $v = \SI{0.5}{\centi\meter}$. To determine regions of interest to investigate, a simple approximation was made to look for spheres of best fit from a distance using MSAC in MATLab (Algorithm \ref{alg:perceptionsphere}). Only rocks with a suitable link-length-to-rock-radius ratio $\ell/r$ were kept as potential regions to extend to. For our specific gripper, $\ell$ was about \SI{60}{\milli\meter}, and there additionally was a palm radius of \SI{30}{\milli\meter}, leading to a full extension of radius \SI{150}{\milli\meter}. A reasonable $\ell/r$ ratio was $0.35$, from a rock with $r = \SI{170}{\milli\meter}$. For the field test, we defined a window $\SI{75}{\milli\meter} \leq r \leq \SI{200}{\milli\meter}$ to quickly filter promising regions of interest.

Once a boom extended to a region, to determine $\alpha$, the raw depth data stream is processed as a point cloud. The point cloud is down-sampled using the same process as above, and normal vectors were estimated from the local neighborhood of points as defined by the $k$ nearest neighbors, $k_n$. The angle $\alpha$ was calculated as the angle between the normal vector and gripper z-axis (Algorithm \ref{alg:perceptionalpha}). Limit surfaces can be generated using the three $\alpha$ values and $r$ value of the grasp in the Monte Carlo simulation.

\paragraph*{Field testing experimental setup}
The deployer was custom-built with three different actuators to provide the necessary functionality of a ReachBot shoulder joint. The deploying mechanism (3D printed, Creality Ender 3 PLA) was a friction-drive wheel powered by a DC motor (Pololu 75:1 Metal Gearmotor). This drive wheel was pressed against the boom coil by compression springs. An additional DC motor (Pololu 75:1 Metal Gearmotor) controlled the elevation of the boom through timing belts and pulleys. The entire assembly was mounted on a turntable (McMaster Carr 6031K16) powered by a servo (Zoskay 35kg) that controlled the pan angle.  

The deployer and gripper were controlled by a single Arduino Mega with manual input from a joystick, and serial commands through USB from a laptop for the gripper. The pan-angle servo was driven using PWM signals from the Arduino Mega. All DC motors were powered by a 12 V power supply and the servo motor was powered by a separate 6 V power supply.

\paragraph*{Statistical Tests}
For the Monte Carlo simulation, the mean of the data is calculated with the sum of all values divided by the total number of values. The standard derivation of the data is calculated by taking the square root of its variance.

\paragraph*{Supplementary Materials}
\textcolor{white}{}\\

Table S1

Figs. S1  to S4

Movies S1

Algorithm pseudocode S1 to S4



\bibliographystyle{Science}
\bibliography{scibib,ASL_papers,main}

\paragraph*{Acknowledgements}
\rev{We thank Stanford University and the Stanford Research Computing Center for providing computational resources and support that contributed to these research results (Sherlock cluster). \textbf{Funding:} Supported by the NASA Innovative Advanced Concepts (NIAC) program. S. Newdick and J. Di are supported by NASA Space Technology Graduate Research Opportunities (NSTGRO). \textbf{Author contributions:} The gripper was designed by T.G.C. The gripper modeling was constructed by T.G.C, J.D. and C.B. The simulation and motion planning was conducted by S.N. The reachbility analysis is performed by N.O. The grasp site detection was conducted by J.D. The field testing were conducted by T.G.C, J.D. and S.N. \textbf{Data and materials availability:} All (other) data needed to evaluate the conclusions in the paper are present in the paper or the Supplementary Materials. Data and code can be found here: 10.5281/zenodo.10836305}

\clearpage
\pagenumbering{arabic}
\renewcommand*{\thepage}{S\arabic{page}}
\renewcommand{\thefigure}{S\arabic{figure}}
\renewcommand{\theequation}{S\arabic{equation}}
\setcounter{section}{0}
\setcounter{figure}{0}
\setcounter{equation}{0}
\renewcommand{\thesection}{S.\arabic{section}}
\renewcommand{\thealgorithm}{S.\arabic{algorithm}}


\titleformat{\subparagraph}
  {\normalfont\rmfamily\itshape\bfseries}{\thesubparagraph}{}{}




\title{Supplementary Materials for
Locomotion as Manipulation with Reachbot}


\author
{Tony G. Chen$^{1\ast}$, Stephanie Newdick$^{2}$, Julia Di$^{1}$, Carlo Bosio$^{1}$, Nitin Ongole$^{2}$,\\ Mathieu Lap\^{o}tre$^{3}$, Marco Pavone$^{2}$, Mark R. Cutkosky$^{1}$\\
\\
\normalsize{$^{1}$Dept. of Mechanical Engineering, $^{2}$Dept. of Aeronautics and Astronautics,}\\
\normalsize{$^{3}$Dept. of Earth and Planetary Sciences, Stanford University,}\\
\normalsize{424 Panama Mall, Stanford 94305, USA}\\
\normalsize{$^\ast$Corresponding author:  agchen@stanford.edu.}
}


\date{}

\maketitle 

\newpage

\section*{Grasp force computation}
\label{sup:gripperforce}

\rev{The details of the derivation of equation \ref{eq:non-slip_3d} are as follows. We create a local coordinate frame at the spine-asperity interface, with a contact plane tangent to the asperity surface. To avoid slip, one can write the following force balance assuming Coulomb friction:}

\begin{equation}
    \sqrt{F_{sx}^{2} + F_{sy}^{2}} \leq \mu F_{sz}.
    \label{eq:non-slip_derivation}
\end{equation}

\begin{figure}[h]
    \centering
    \includegraphics[width=0.75\textwidth]{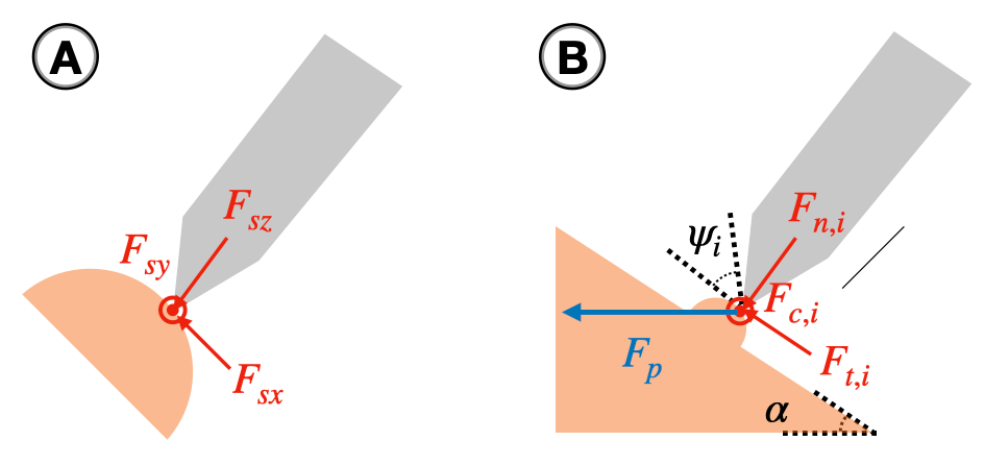}
    \caption{\textbf{Spine-level contact modeling.} \rev{(A) Coulomb friction limit at spine contact. (B) A single spine interacting on an asperity with angle $\psi_i$, with its force decomposed into components.}}
    \label{fig:eq7supple}
\end{figure}

\rev{We decompose the spine-asperity interface force into four different components: $F_{n,i},F_{c,i},F_{t,i}$, which are the normal and tangential components, and $F_{p,i}$, which is the internal force applied by the gripper. We then construct the following rotation matrix to transform the four different components of the interaction force into the local coordinate frame:}

\begin{equation}
    \begin{bmatrix}
        \cos\psi_i & -\sin\psi & 0 & \cos(\psi + \alpha) \\
        0 & 0 & 1 & 0 \\
        \sin(\psi)& \cos(\psi) & 0 & \sin(\psi + \alpha)
    \end{bmatrix}
    \begin{bmatrix}
        F_{t,i} \\
        F_{n,i} \\
        F_{c,i} \\
        F_{p,i}
    \end{bmatrix}
    =
    \begin{bmatrix}
        F_{sx} \\
        F_{sy} \\
        F_{sz} \\
    \end{bmatrix}
\end{equation}

In equation (\ref{eq:3d_symm}) the details of the components are:

\begin{equation}
    \mathbf{A} = \begin{bmatrix}
             \text{s}_\alpha & \text{c}_\alpha & 0 & -\frac{1}{2}\text{s}_\alpha & -\frac{1}{2}\text{c}_\alpha & -\frac{\sqrt{3}}{2} & -\frac{1}{2}\text{s}_\alpha & -\frac{1}{2}\text{c}_\alpha & \frac{\sqrt{3}}{2} \\
             0 & 0 & 1 & \frac{\sqrt{3}}{2}\text{s}_\alpha & \frac{\sqrt{3}}{2}\text{c}_\alpha & -\frac{1}{2} & -\frac{\sqrt{3}}{2}\text{s}_\alpha & -\frac{\sqrt{3}}{2}\text{c}_\alpha & -\frac{1}{2} \\
             \text{c}_\alpha & -\text{s}_\alpha & 0 & \text{c}_\alpha & -\text{s}_\alpha & 0 & \text{c}_\alpha & -\text{s}_\alpha & 0 \\
             0 & 0 & 0 & \frac{\sqrt{3}}{2}\text{s}_\alpha\text{c}_\alpha & -\frac{\sqrt{3}}{2}\text{s}_\alpha^2 & 0 & -\frac{\sqrt{3}}{2}\text{s}_\alpha\text{c}_\alpha & \frac{\sqrt{3}}{2}\text{s}_\alpha^2 & 0 \\
             -\text{s}_\alpha\text{c}_\alpha & \text{s}_\alpha^2 & 0 & \frac{1}{2}\text{s}_\alpha\text{c}_\alpha & -\frac{1}{2}\text{s}_\alpha^2 & 0 & \frac{1}{2}\text{s}_\alpha\text{c}_\alpha & -\frac{1}{2}\text{s}_\alpha^2 & 0\\
             0 & 0 & 1 & 0 & 0 & 1 & 0 & 0 & 1 \\
             \frac{\text{s}_\alpha}{k_n} & \frac{\text{c}_\alpha}{k_t} & 0 & \frac{\text{s}_\alpha}{k_n} & \frac{\text{c}_\alpha}{k_t} & 0 & \frac{\text{s}_\alpha}{k_n} & \frac{\text{c}_\alpha}{k_t} & 0 \\
             \frac{\text{s}_\alpha}{k_n} & \frac{\text{c}_\alpha}{k_t} & 0 & 0 & 0 & \frac{1}{\sqrt{3}k_c} & 0 & 0 & -\frac{1}{\sqrt{3}k_c} \\
             \frac{\text{s}_\alpha}{2 k_n} & \frac{\text{c}_\alpha}{2 k_t} & \frac{\sqrt{3}}{2 k_c} & 0 & 0 & 0 & \frac{\text{s}_\alpha}{k_n} & \frac{\text{c}_\alpha}{k_t} & 0 
        \end{bmatrix}
\end{equation}

\begin{equation}
        F = \begin{bmatrix}
             F_n^1, F_t^1, F_c^1, F_n^2, F_t^2, F_c^2, F_n^3, F_t^3, F_c^3
        \end{bmatrix}^T
\end{equation}

\begin{equation}
        B = \begin{bmatrix}
             -F_{pull}\text{s}_\beta\text{c}_\phi \\ -F_{pull}\text{s}_\beta\text{s}_\phi \\ -F_{pull}\text{c}_\beta \\ F_{pull}\text{s}_\beta\text{s}_\phi(1-\text{c}_\alpha + \frac{m}{r}) \\ -F_{pull}\text{s}_\beta\text{c}_\phi(1-\text{c}_\alpha + \frac{m}{r}) \\ 0 \\ 0 \\ 0 \\ 0
        \end{bmatrix}
\end{equation}

where, for compactness, $c_\alpha = \cos{\alpha}$, etc. 
These matrices are derived from (i) equilibrium equations applied to the gripper (the first six rows) and (ii) kinematic conditions derived from the motion of a rigid body applied to the grasped rock (satisfied if the fingers do not lose contact).
With reference to Fig. \ref{fig:3d_eq}, the equilibrium equations are written with respect to a reference frame aligned with the global frame and centered in the point $G$, i.e. the centroid of the three contact points. Rows 1 to 3 of the matrix system represent the force equilibrium, while rows 4 to 6 represent the moment equilibrium. The resultant moment along the $z$ axis is zero because the direction of the pulling force, applied from the boom to the wrist joint, and the $z$ axis always intersect.

\begin{figure}[h]
    \centering
    \includegraphics[width=0.4\textwidth]{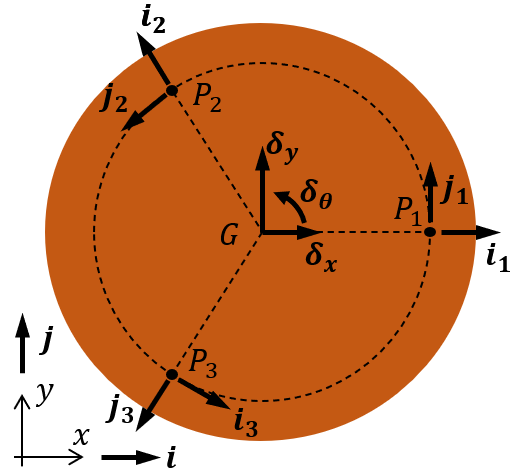}
    \caption{\textbf{Three-point contact on a rock with the gripper.} The three contact points from the gripper determine a plane within the rock.}
    \label{fig:displ}
\end{figure}

Rows 7 to 9 arise from kinematic relations, cast into contact force relations through the compliance terms introduced, involving fingertips displacements. The three contact points determine a plane within the rock, as shown in Fig.~\ref{fig:displ}, with the contact points $P_i$ ($j = 1, 2, 3$). The displacements of the three contact points $\Delta_j$  can be written in terms of the overall kinematics of the rock ($\delta_x, \delta_y, \delta_\theta$) and projected along the local frames aligned with the components of the contact forces (${x_1, y_1, z_1}$). In particular we have:
\begin{equation}
        \begin{cases}
            \Delta_1 = \delta_x \mathbf{i_1} + (\delta_y + r' \delta_\theta) \mathbf{j_1} \\
            \Delta_2 = (-\frac{1}{2} \delta_x + \frac{\sqrt{3}}{2}\delta_y)\mathbf{i_2} + (-\frac{\sqrt{3}}{2}\delta_x - \frac{1}{2} \delta_y + r' \delta_\theta) \mathbf{j_2} \\
            \Delta_3 = (-\frac{1}{2} \delta_x - \frac{\sqrt{3}}{2}\delta_y)\mathbf{i_3} + (\frac{\sqrt{3}}{2}\delta_x - \frac{1}{2} \delta_y + r' \delta_\theta) \mathbf{j_3}
        \end{cases}
    \end{equation}
where $r' = r \sin \alpha$. The components of the three displacement vectors of the contact points are directly related to the contact forces. In fact, the displacements along the local unity vectors $\mathbf{i_i}$ and $\mathbf{j_i}$ ($i = 1,2,3$) can be expressed in terms of ratios between contact forces and spine compliances. In particular, by substituting, the following system of equations is obtained:

\begin{equation}
        \begin{cases}
            -\delta_x = \frac{\sin \alpha}{k_n} F_n^1 + \frac{\cos \alpha}{k_t} F_t^1 \\
            -\delta_y - r' \delta_\theta = \frac{1}{k_c} F_c^1 \\
            \frac{1}{2} \delta_x - \frac{\sqrt{3}}{2} \delta_y = \frac{\sin\alpha}{k_n} F_n^2 + \frac{\cos\alpha}{k_t} F_t^2 \\
            \frac{\sqrt{3}}{2} \delta_x + \frac{1}{2} \delta_y - r' \delta_\theta = \frac{1}{k_c} F_c^2 \\
            \frac{1}{2} \delta_x + \frac{\sqrt{3}}{2} \delta_y = \frac{\sin\alpha}{k_n} F_n^3 + \frac{\cos\alpha}{k_t} F_t^3 \\
            -\frac{\sqrt{3}}{2}\delta_x + \frac{1}{2} \delta_y - r'\delta_\theta = \frac{1}{k_c}F_c^3
        \end{cases}
        \label{eq:constraints}
    \end{equation}
If we manipulate this system, we get three equations in which the dependence on the kinematics of the rock disappear. These represent kinematic constraints between contact forces at the fingertips which need to be satisfied so that the spines maintain contact and do not penetrate the rock.

\section*{Monte Carlo limit surface simulation setup \label{sup:MC}}

The Monte Carlo sampling of the limit surface was set up starting from the grasp model introduced 
for equations \ref{eq:non-slip_3d} and \ref{eq:3d_symm}.
We sweep the angles defining the direction of the force pulling on the wrist joint of the gripper ($\beta$ and $\phi$ with reference to Fig.~\ref{fig:3d_eq}). In particular, the continuous set $[0,\pi/2]\times [0,2\pi/3]$ where the tuple $(\beta, \phi)$ lives is discretized in a $n_\text{grid}\times n_\text{grid}$ grid. 
For each point of the grid, we sample $n_\text{MC}$ sets of asperity angles $\psi_i$ ($i = 1, ..., 3 n_{spines}$ where $n_{spines}$ is the number of spines per finger). The asperities angle distribution is assumed to be approximated by a uniform distribution in the form of $U(0,\pi/2)$. This is justified by measurements presented in \cite{asbeck2006scaling} for very rough surfaces as concrete and cobblestone. For each of these samples, we gradually increase the magnitude of the pulling force until grasp failure is reached, which can happen in two ways:
(i) all the spines on one finger fail or (ii) a whole finger loses contact. Whenever a spine fails (this could happen because of slip or because the load on a spine overcomes a strength threshold $F_\text{max}$), the load redistributes on the remaining active spines. This can cause other spines on the same finger to fail. \rev{We do not resample the asperity distribution at this time.}
Once all the maximum pulling force magnitudes are computed, possible to approximate through samples the underlying distribution. In particular, it is possible to compute the mean, or any other statistic of relevance for the problem, as for example a certain percentile which guarantees a high enough confidence on the grasp performances.

We provide the pseudocode (algorithm \ref{alg:MC}).
\begin{algorithm}
\caption{Monte Carlo sampling of Limit Surface}\label{alg:MC}
\begin{algorithmic}
    \Require $\text{Parameters} \,\, \ell, \alpha, r, x, n_{spines}, F_p, k_n, k_t, k_c, F_\text{max}$
    \Require $\text{Hyperparameters} \,\, n_\text{grid}, n_\text{MC}$
    \State Discretize $(\beta, \phi)$ support
    \For{every grid point $(\beta, \phi)$}
    \State Sample $n_\text{MC}$ sets of asperity angles
    \For{every sample}
    \State Set pulling force to zero
    \While{an entire finger has not failed}
    \State Increase pulling force
    \State Compute fingertips contact forces
    \State Check no-slip condition at every spine
    \If{a spine failed}
    \State Set the spine as inactive
    \State Redistribute load on remaining spines of that finger
    \EndIf
    \State Check failure of entire finger
    \EndWhile
    \State Store value of max pulling force magnitude 
    \EndFor
    \State Compute max pulling force statistics for current grid point
    \EndFor
\end{algorithmic}
\end{algorithm}

\clearpage
\section*{Gripper engineering parameters}
\label{sup:gripper_parameters}

\setcounter{table}{0}

\begin{table}[h]
    \centering
    \caption{Gripper engineering parameters for ReachBot}
    \begin{tabular}{||c | c ||}
        \hline
        gripper mass & \SI{290}{\gram} \\
        \hline
        palm dimension (LxWxH) & \SI{80}{\milli\meter} x \SI{63}{\milli\meter} x \SI{36}{\milli\meter} \\
        \hline
        distal phalange length & \SI{53}{\milli\meter} \\
        \hline
        proximal phalange length & \SI{60}{\milli\meter}\\
        \hline
        proximal distal magnet strength & \SI{2}{\newton}  \\
        \hline
        spine tile size (LxWxH) & \SI{30}{\milli\meter} x \SI{16.5}{\milli\meter} x \SI{19}{\milli\meter} \\
        \hline
        number of spines per tile & 40\\
        \hline
        spine tile linear travel & \SI{10}{\milli\meter}\\
        \hline
        spine tile spring stiffness & \SI{4.5}{\newton\per\centi\meter}\\
        \hline
        spine tile torsional travel & \SI{-10}{\degree} to \SI{30}{\degree}\\
        \hline
        spine normal travel & \SI{8}{\milli\meter}\\
        \hline
        spine normal spring stiffness & \SI{0.2627}{\newton\per\centi\meter}\\
        \hline
    \end{tabular}
    \vspace{6pt}
    \label{parameters}
\end{table}
\vspace{-4mm}

\section*{Geometric relations among grasp parameters}
\label{sup:grasp_elltoalpha}

With reference to Fig. \ref{fig:3d_eq}, it is possible to obtain some geometric relations between grasp parameters. In particular, we can write the geometric closure of the system:
\begin{equation}
\begin{cases}
    \ell \sin \alpha = r(1 - \cos \alpha) \\
    \ell(1 + \cos \alpha) = r \sin \alpha
\end{cases}
\label{eq:grasp_elltoalpha}
\end{equation}

In which we considered a simplified case when $x=0$.
This leads to an approximate, but simple, relation between the geometric parameters of interest. It is possible to make the dependency of $\alpha$ on the term $\ell/r$ explicit:

\begin{equation}
    \cos \alpha = \frac{1 - (\ell/r)^2}{1 + (\ell/r)^2}
\end{equation}

This is related to the analytic expression used in the plot of Fig. \ref{fig:perception}(C).

\section*{\rev{Mojave volcanic rock surface scan}}
\label{sup:mojave_surface}

\rev{Surface waviness at scale of \SIrange{1}{2}{\milli\meter} up to \SI{2}{\centi\meter} determines how much linear travel the spines will need within each tile. To address this question field tests were conducted on volcanic rock from the Lavic Lake volcanic field in Mojave. After field testing was completed, the tested rock surfaces were scanned using a Creality CR-SCAN 01 structured light scanner in handheld mode. The Creality CR-SCAN 01 reports accuracy up to \SI{0.1}{\milli\meter} and a scan resolution up to \SI{0.5}{\milli\meter}. We note, however, that these specifications are likely for stationary mode and that handheld mode will be a little worse.}

\begin{figure}[h]
    \centering
    \includegraphics[width=0.6\textwidth]{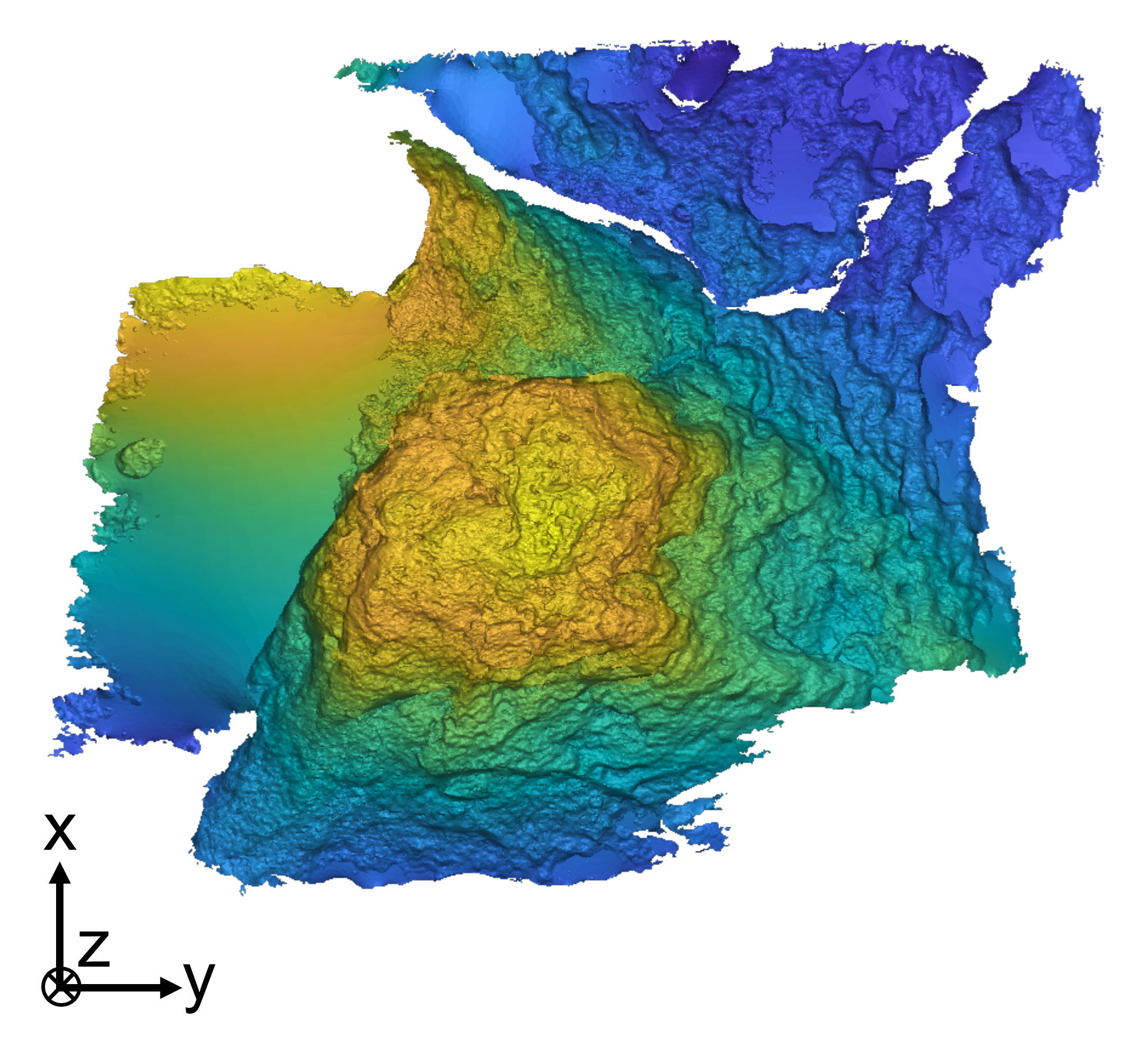}
    \caption{\textbf{Scanned rock.} Scan of the grasped rock feature shown in Fig. \ref{fig:glamour}B-C.}
    \label{fig:supp_scan}
\end{figure}

\begin{figure}[h]
    \centering
    \includegraphics[width=0.9\textwidth]{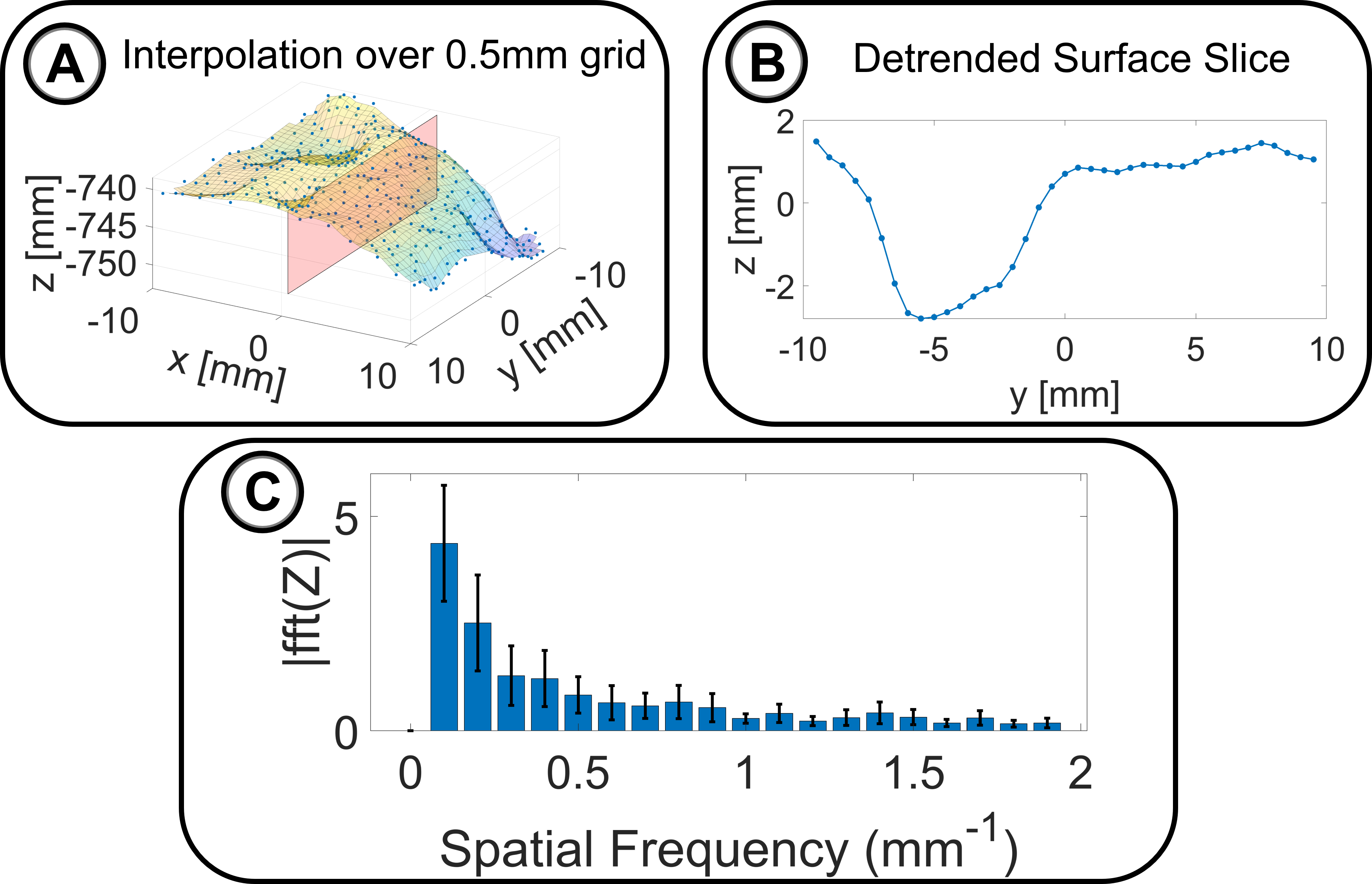}
    \caption{\textbf{Rock surface analysis.} (A) The scanned surface at the scale of a microspine tile (2x2\,cm). The red plane is a cutting plane. (B) The surface profile that a single spine would encounter along the red cutting plane. (C) A Fast Fourier Transform taken of 5 different scanned rock surfaces shows a spike at a centimeter frequency which tapers off for frequencies down to the millimeter scale.}
    \label{fig:supp_mojavefft}
\end{figure}

\rev{Given a scan of a rock feature, such as that shown in Fig.~\ref{fig:supp_scan}, we can obtain a surface patch the size of a spine tile (2 x \SI{2}{\centi\meter}) to determine surface characteristics. The Creality scan returns non-uniform datapoints, so we interpolate to uniformly distribute the data across a \SI{0.5}{\milli\meter} grid, which was chosen based on the manufacturer specification (Fig.~\ref{fig:supp_mojavefft}(A)). We then take a slice from the surface to obtain the surface profile that a spine follows when contacting the surface, and de-trend this profile (Fig.~\ref{fig:supp_mojavefft}(B)). We use a Fast Fourier Transform to detect a characteristic spatial wavelength, and repeat this process for 5 different rock surfaces from the cave and report these statistics (Fig.~\ref{fig:supp_mojavefft}(C)). Although we did not comprehensively scan 
surfaces in the lava tubes, based on a qualitative judgement of the rock surfaces from field testing, we believe that the captured profiles are representative.}

\rev{As seen in Fig.~\ref{fig:supp_mojavefft}(C), there is a centimeter-scale surface waviness present in the tested rocks. As noted in the gripper design section, the 10\,mm linear spine travel accommodates this waviness. For surfaces with different waviness, we may accordingly scale ReachBot's spine travel, tile size and spine-to-spine spacing.}

\section*{Pseudocode for grasp reachability}

We used the following algorithm for conducting the grasp reachability analysis.

\begin{algorithm}[H]
\caption{Grasp reachability}\label{alg:reachability}
\begin{algorithmic}
    \Require Finger geometric parameters and constraints
    \Require Triangulated rock mesh, face outward normals
    \Require Hyperparameter: sampling fidelity
    \For{all cases considered}
    \State Sample possible states of the finger for the case considered
    \For{every sample}
    \If{the finger does not collide with a face}
    \State Store the sample as a good sample
    \EndIf
    \EndFor
    \For{every face}
    \For{every good sample}
    \If{the spine intersects the face at a $10^\circ{}-30^\circ{}$ angle from its normal}
    \State Store the face as a good face
    \State Break
    \EndIf
    \EndFor
    \EndFor
    \EndFor
\end{algorithmic}
\end{algorithm}

\section*{Pseudocode for perception field test}

We used the following algorithms for conducting the sphere fitting and alpha estimation during the field tests.

\begin{algorithm}
\caption{Sphere Fitting}\label{alg:perceptionsphere}
\begin{algorithmic}
    \Require Input RGBD pointcloud
    \Require $\text{Parameters} \,\, v, d_{max}, 
    \ell, r_{min}, r_{max}$
    \State Downsample raw pointcloud using voxel size $v$
    \For{every voxel point}
    \State Return average of voxels within size $v$
    \EndFor
    \For{every downsampled voxel}
    \State Set maximum distance to be considered an inlier for sphere model $d_{max}$
    \State Compute sphere fit with MSAC
    \If{Computed model radius is not within desired $\ell/r$ window} 
    \State Discard model
    \EndIf
    \EndFor
\end{algorithmic}
\end{algorithm}

\begin{algorithm}
\caption{Alpha Estimation}\label{alg:perceptionalpha}
\begin{algorithmic}
    \Require Input RGBD pointcloud from desired sphere region
    \Require $\text{Parameters} \,\, v, k_n, 
    \ell, \alpha_{min}, \alpha_{max}$
    \State Downsample raw pointcloud using voxel size $v$
    \For{every voxel point}
    \State Return average of voxels within size $v$
    \EndFor
    \For{every downsampled voxel}
    \State Define the normal vector with a plane fitted using the $k_n$ nearest neighbors
    \State Let $\alpha$ be the angle between the estimated normal and the z-axis
    \If{$\alpha$ is within $\alpha_{min}$ to $\alpha_{max}$ degrees}
    \State Set area as graspable
    \Else{ Set area as un-graspable}
    \EndIf
    \EndFor
\end{algorithmic}
\end{algorithm}

\section*{Movie of field test in lava tube}

\textbf{Movie of field test in lava tube.} The attached Movie S1 showcases ReachBot technology in an unnamed lava tube near Pisgah Crater. The ReachBot single-arm prototype was able to target, deploy, and grasp many target sites within the lava tube, with one full sequence shown here.


\end{document}